\documentclass{article}

\usepackage{arxiv}

\usepackage[utf8]{inputenc} 
\usepackage[T1]{fontenc}    
\usepackage{hyperref}       
\usepackage{url}            
\usepackage{booktabs}       
\usepackage{amsfonts}       
\usepackage{nicefrac}       
\usepackage{microtype}      
\usepackage{lipsum}
\usepackage{authblk}

\usepackage[numbers]{natbib}
\usepackage{float} 
\usepackage{times}
\usepackage{epsfig}
\usepackage{graphicx}
\usepackage{amsmath, amsfonts, amssymb}
\usepackage{bm}
\usepackage{algorithm}
\usepackage{algorithmicx}
\usepackage{comment}
\usepackage{enumitem}
\usepackage{fullpage}
\usepackage{tikz}
\usepackage{pifont}
\usepackage{amsthm}
\usepackage{subcaption}
\newcommand{\cmark}{{\color{blue}\ding{51}}}%
\newcommand{\xmark}{{\color{red}\ding{55}}}

\usepackage{amsmath,amsfonts,amssymb}
\usepackage{mathtools}
\usepackage{multirow}
\usepackage[noend]{algpseudocode}
\usepackage{booktabs}
\usepackage[export]{adjustbox}
\usepackage{url}

\newtheorem*{remark}{Remark}

\usepackage{xspace}

\newcommand*{\ie}{\textit{i.e.}\@\xspace}
\newcommand*{\st}{\textit{s.t.}\@\xspace}
\newcommand*{\vs}{\textit{vs.}\@\xspace}
\newcommand*{\wrt}{\textit{w.r.t.}\@\xspace}
\makeatletter
\newcommand*{\etc}{%
	\@ifnextchar{.}%
	{\textit{etc}}%
	{\textit{etc.}\@\xspace}%
}
\makeatother
\algtext*{EndWhile}
\algtext*{EndFor}
\algtext*{EndIf}
\algdef{SE}[DOWHILE]{Do}{doWhile}{\algorithmicdo}[1]{\algorithmicwhile\ #1}%

\makeatletter
\def\BState{\State\hskip-\ALG@thistlm}
\makeatother

\title{SCA-Net: A Self-Correcting Two-Layer Autoencoder for Hyper-spectral Unmixing}

\author[1]{\textbf{Gurpreet Singh} \textsuperscript{\dag}}
\author[1,2]{\textbf{Soumyajit Gupta} \textsuperscript{\dag}}
\author[1,3]{\textbf{Clint Dawson}}
\affil[2]{Department of Computer Science}
\affil[3]{Oden Institute for Computational Engineering and Sciences}
\affil[1]{The University of Texas at Austin}
\affil[ ]{\texttt{\{gurpreet, smjtgupta\}@utexas.edu}, \texttt{clint.dawson@oden.utexas.edu}}

\begin{document}

\maketitle

{\let\thefootnote\relax\footnote{{\dag contributed equally to this work.}}}

\begin{abstract}
Hyperspectral unmixing involves separating a pixel as a weighted combination of its constituent endmembers and corresponding fractional abundances, with the current state of the art results achieved by neural models on benchmark datasets. However, these networks are severely over-parameterized and consequently, the invariant endmember spectra extracted as decoder weights have a high variance over multiple runs. These approaches perform substantial post-processing while requiring an exact specification of the number of endmembers and specialized initialization of weights from other algorithms like VCA. We show for the first time that a two-layer autoencoder (SCA), with $2FK$ parameters ($F$ features, $K$ endmembers), achieves error metrics that are scales apart ($10^{-5})$ from previously reported values $(10^{-2})$. SCA converges to this low error solution starting from a random initialization of weights. We also show that SCA, based upon a bi-orthogonal representation, performs a self-correction when the number of endmembers are over-specified. Numerical experiments on Samson, Jasper, and Urban datasets demonstrate that SCA outperforms previously reported error metrics for all the cases while being robust to noise and outliers. 
\end{abstract}

\section{Introduction}

Hyperspectral Image (HSI) datasets capture spatial and spectral information for image analysis applications including classification and unmixing, in fields such as agriculture, environment, mineral mapping, surveillance, and chemical imaging \cite{bioucas2013hyperspectral}. However, due to the relatively low spatial resolution, pixels carry information of a mixture of several materials complicating the separation and characterization of such datasets for practical use. Hyperspectral Unmixing addresses this problem by decomposing each pixel spectrum into a set of pure spectra (\ie endmembers) and their corresponding proportions (\ie abundances) as a Linear Mixture Model (LMM).

A number of algorithms have been proposed for HSU \cite{bioucas2013hyperspectral,zhu2017hyperspectral} assuming different mixture models that can be broadly categorized into two classes: linear \cite{bioucas2013hyperspectral} and nonlinear \cite{dobigeon2013nonlinear} mixture models. Linear Mixing Model (LMM) is valid when the mixing scale is macroscopic and the photon reaching the sensor has interacted with just one material. In contrast, Non-Linear Mixture Model considers physical interactions (interference, refraction, \etc) between light scattered by multiple materials at a microscopic level.

We now discuss LMM as a starting point for our proposed solution. Let the HSI data matrix be $Y_F \in \mathbb{R}_+^{N \times F}$, having non-negative entries, where $F$ is the feature/spectral length and $N$ is the number of samples/pixels in the data. As per LMM, the spectrum of each pixel $y_F \in \mathbb{R}^F$ is an additive mixture of endmembers with fractional abundances given by,
\begin{align} 
Y_F = A E + \gamma \quad \mathrm{\textit{s.t.}} \quad a_{i,k} \geq \mathbf{0}, \sum_{k=1}^{K}a_{i,k}=1 \label{eq:lmm}
\end{align}
where $K \leq F$, the matrix $E=[e_1,\ldots,e_K]^T \in \mathbb{R}_+^{K \times F}$ of endmembers, $A=[a_1,\ldots,a_K] \in (\mathbb{R}_{+} \cup \{0\})^{N \times K}$ of per-pixel abundance, and $\gamma$ as additive perturbation (noise and modeling errors). By definition, at most $K$ endmembers (materials) are present in the HSI data, where the endmembers form a non-orthogonal basis spanning a rank-K sub-space of $Y_F$. Additionally, an LMM requires that the extracted endmembers $E$ and abundances $A$ satisfy the following constraints:
\begin{enumerate}[leftmargin=*]
	\item \textbf{Abundance Non-negativity Constraint (ANC):} $a_{i,k} \geq 0, \forall i,k$, materials contribute additively.
	\item \textbf{Abundance Sum Constraint (ASC):} $A.\mathbf{1}_{K}=\mathbf{1}_{N}$ to represent percentage presence of a material.
	\item \textbf{Non-negative, bounded spectral values:} $0 \leq e_{k,j} \leq 1$ assumes HSI end members are strictly positive.
\end{enumerate}


We find a number of limitations in the current state of the art methods: \textbf{(1)} Approaches that rely upon dividing the problem into separate endmember extraction followed by abundance estimation assume that, once an approximation to endmembers $E$ is extracted, estimation of $A$ can be carried out independently. However, note that, given a dataset $Y_F$, LMM requires $Y_F = AE$. Since $Y_F$ does not change, this implies any estimation of $A$ as an independent post-processing step will need to alter $E$ to abide by $Y_F=AE$. This holds true for both neural \cite{ozkan2018endnet} and non-neural \cite{nascimento2005vertex} approaches if the post-processing algorithm is not a linear operation. \textit{In contrast, SCA extracts both the endmembers and abundances by solving a single minimization problem where both pre and post-processing steps are simple linear scaling and un-scaling operations. Note that abundances are directly obtained from SCA without an independent post-processing step.}

\textbf{(2)} Prior approaches rely upon initialization from other algorithms (VCA \cite{nascimento2005vertex}) to obtain better approximations. Specifically, the current state of the art Endnet \cite{ozkan2018endnet} does not clarify how the network weights are initialized making it difficult to reproduce their results. For a network to be initialized properly: \textbf{a)} the network weights must be fully interpretable and \textbf{b)} the number of network parameters must match the size of the initialization vector obtained from other algorithms. \textit{Although not required, SCA can be trivially initialized by approximations of $E$ from other algorithms since both the encoder and decoder weights have a distinct meaning.}  

\textbf{(3)} Neural approaches \cite{ozkan2018endnet,su2019daen,khajehrayeni2020hyperspectral} show a high variance in extracted endmembers even when the dataset $Y_F$ is static. This is due to an over-parametrized network specification wherein SCA has exactly $2FK$ parameters. For non-neural approaches such as NMF \cite{qian2011hyperspectral,zhu2014spectral} both endmembers $E\in \mathbb{R}_{+}^{K\times F}$ and abundances $A \in(\mathbb{R}_{+} \cup \{0\})^{N \times K}$ are unknowns. The number of parameters for these approaches (including pre and post-processing steps) are at least $K(F+N)$ and therefore dependent on the number of samples, leading to scalability issues as sample size increases. SCA network parameters ($2FK$), as with other neural approaches, are sample independent. \textbf{Table \ref{tab:comp}} presents a brief comparison of SCA with existing approaches in the light of the aforementioned limitations.
\begin{table}[ht]
    \centering
    \begin{tabular}{c|ccccc|c}
    \toprule
        Method & GAEEII & EndNet & VCA & $l_{1|2}$-NMF & DgS-NMF & \bf SCA \\ \midrule
        Abides LMM & \xmark & \xmark & \xmark & \cmark & \cmark & \cmark \\
        Interpretable & \xmark & \xmark & \cmark & \cmark & \cmark & \cmark \\
        Scalable & \cmark & \cmark & \cmark & \xmark & \xmark & \cmark \\
        Random Init. & \xmark & \xmark & \cmark & \xmark & \xmark & \cmark \\ \bottomrule
    \end{tabular}
    \caption{SCA \vs existing state-of-the-art methods.}
    \label{tab:comp}
    \vspace{-2mm}
\end{table}

\noindent {\bf Contributions.} Our key contributions are as follows:
\begin{enumerate}[leftmargin=*]
    \item SCA is low weight ($2FK$) and fully interpretable autoencoder where all the network weights and outputs have a specific meaning.
    \item A bi-orthogonal representation renders SCA a self-correcting property for over-specified endmembers.
    \item SCA consistently converges to a low error solution with random weights initialization over multiple runs.
    \item The network loss is bounded below by a computationally verifiable tail energy following Eckart-Young-Mirsky (EYM) theorem.
    \item SCA formulation is robust to noise and outliers.
\end{enumerate}

\section{Related Works} \label{sec:related}

LMM solution strategies fall under three categories: 1. {\em Supervised}:  endmembers are known a priori, where they are extracted from the data via endmember extraction algorithms \cite{nascimento2005vertex} or captured from spectral libraries \cite{aster}. 2. {\em Semi-supervised}: the optimal subset of endmembers that suits the data are estimated from spectral libraries in advance \cite{themelis2010semi}. {\em Unsupervised}: both endmembers and the corresponding abundances are estimated simultaneously from HSIs, given the number of endmembers \cite{chan2011simplex}. 


\noindent \textbf{Pure Pixel Approaches:} These methods work under the assumption that the dataset contains at least one sample corresponding to each of the endmembers. PPI \cite{boardman1995mapping}, N-FINDR \cite{winter1999n} and VCA \cite{nascimento2005vertex} fall under this category. The common thread across these methods is to find a  projection of the HSI data such that maximal information is preserved that satisfy a volume criterion or extract orthogonal features. Extensions include IEA \cite{neville1999automatic}, SGA \cite{chang2006new}, SMACC \cite{gruninger2004sequential}, SVMAX \cite{chan2011simplex} \etc. Once the endmembers are estimated, least-squares based approaches, like FCLS \cite{heinz2001fully}, are used to estimate the abundances. 

\noindent \textbf{Minimum Volume based approaches}: These methods work under the assumption the data samples span the volume captured by the endmembers. The objective is to find a mixing matrix that minimizes the volume of the simplex defined by its columns, such that the simplex encloses the observed spectral vectors. Readers are referred to \cite{bioucas2013hyperspectral} for a detailed description. Methods under this category include MVSA \cite{li2008minimum}, MVES \cite{chan2009convex}, ICE \cite{berman2004ice} and CCA \cite{ifarraguerri1999multispectral}. All these methods have variations across the volume criteria they operate on and additional penalization placed on the estimated endmembers.

\noindent \textbf{Statistical approaches}: These include variants of Non-negative Matrix Factorization (NMF) since the LMM requires that both of its factors are element-wise positive. Since a non-convex optimization for NMF fails to ensure a unique solution, these methods rely on explicit initialization by techniques like VCA. These methods also work when the data points do not span the entire volume of the endmember simplex. Readers are referred to \cite{zhu2017hyperspectral} for a detailed description. Variants include MVCNMF \cite{miao2007endmember}, GNMF \cite{cai2010graph}, DgS-NMF \cite{zhu2014spectral} and $l_{1|2}$-NMF \cite{qian2011hyperspectral}.

\noindent \textbf{Neural approaches}: All works under neural setting follow an autoencoder approach. The idea is to reconstruct the input data at the decoder end and enforce loss functions on the encoder output forcing it to learn the end-members. They rely on explicit initialization of decoder weights by methods like VCA, as they cannot arrive at the solution under random weight setting. Examples include DAEN \cite{su2019daen}, DCAE \cite{khajehrayeni2020hyperspectral} and EndNet \cite{ozkan2018endnet}.

Given extensive prior literature, we refer readers to the survey papers \cite{bioucas2013hyperspectral,zhu2017hyperspectral} and the citations therein for a detailed description. Our review  of prior works shows that neural and genetic approaches have the best reported error metrics. We therefore refer to the values presented in GAEEII \cite{soares2019gaeeii}, DCAE \cite{khajehrayeni2020hyperspectral} and EndNet \cite{ozkan2018endnet} as the best ones. \textbf{Table \ref{tab:comp-soa}} shows a comparison of these current best methods using SAD and RMSE error metrics. One can observe that Endnet error metrics show them to be state of the art and \textit{till date we have not found better reported error values across the three datasets considered in this work.} Note that the error values for SAD and RMSE are at a scale of $10^{-2}$ in prior works.

\begin{table}[ht]
    \centering
    \begin{tabular}{c|ccccc}
    \toprule
         & \multicolumn{5}{c}{Spectral Angle Distance (SAD) $(\times 10^{-2})$} \\
        Method & VCA & $l_{1|2}$-NMF & DgS-NMF & GAEEII & EndNet \\ \midrule
        Samson & 13.17$\pm$1.0 & 7.80$\pm$3.2 & 5.05$\pm$2.7 & 3.54$\pm$1.16 & \textbf{2.98$\pm$0.2} \\
        Jasper & 33.73$\pm$6.2 & 7.19$\pm$2.4 & 5.41$\pm$0.1 & 5.87$\pm$2.65 & \textbf{3.91$\pm$0.5}\\
        Urban & 41.77$\pm$4.5 & 11.01$\pm$0.2 & 8.55$\pm$0.2 & 6.49$\pm$2.34 & \textbf{4.42$\pm$0.3} \\ 
    \bottomrule
    \toprule
         & \multicolumn{5}{c}{Root Mean Square Error (RMSE) $(\times 10^{-2})$} \\
        Method & VCA & $l_{1|2}$-NMF & DgS-NMF & GAEEII & EndNet \\ \midrule
        Samson & 19.66$\pm$3.2 & 7.1$\pm$2.4 & 6.07$\pm$2.8 & 1.95$\pm$0.67 & {3.88$\pm$0.0} \\
        Jasper & 12.65$\pm$4.1 & 11.37$\pm$0.2 & 8.15$\pm$0.2 & 4.04$\pm$2.58 & {7.96$\pm$0.3} \\
        Urban & 30.79$\pm$4.7 & 12.62$\pm$0.1 & 10.49$\pm$0.1 & 4.07$\pm$1.53 & {9.23$\pm$0.2} \\ \bottomrule
    \end{tabular}
    \caption{Error Metrics of Current Best Methods. DCAE's definition is wrong for RMSE, hence we are not reporting its numbers. EndNet performs the best, but stuck at $10^{-2}$ for both SAD and RMSE.}
    \label{tab:comp-soa}
    \vspace{-5mm}
\end{table}

\section{Self Correcting Autoencoder (SCA)}

Given data $Y_F \in \mathbb{R}_+^{N \times F}$, with $K$ endmembers, any autoencoder design needs to construct encoder $\tilde{E} \in \mathbb{R}^{F \times K}$ and decoder $E \in \mathbb{R}^{K \times F}$ weights \st the reconstructed data $\tilde{Y}_F$ is close to $Y_F$ under an appropriate norm.
\begin{align}
    Y_F \tilde{E} E = (AE)\tilde{E} E = A(E\tilde{E}) E = \tilde{Y}_F \label{eq:form}
\end{align}
Ideally, one would like $Y_F$ to be equal to $\tilde{Y}_F$. This is true \textit{iff} $E\tilde{E}=\mathbf{I}^{K}$ in Eq. \ref{eq:form}. However, one can see that more than one solution can exist \st $E\tilde{E}=\mathbf{I}^{K}$. We therefore impose additional requirements on this autoencoder structure so that it explicitly provides us with the endmember spectra $E$ and their corresponding abundances $A$. This implies: 
\begin{align}
    Y_F\tilde{E} = A \label{eq:abun}
\end{align}
This represents the action of the encoder weights $\tilde{E}$ on the data $Y_F$ resulting in the encoder output $A$ as fractional abundances.
One can easily see that multiplying both sides of Eq. \ref{eq:abun} with $E$ results in the LMM formulation in Eq. \ref{eq:lmm} and consequently the decoder output $\tilde{Y}_F$ in Eq. \ref{eq:form}. Mathematically, $E\tilde{E}=\mathbf{I}^K$ is a discrete bi-orthogonal representation \cite{andrle2007experiments}, different from the well known orthogonal representation in Singular Value Decomposition (SVD).

\begin{remark}
    The decoder matrix $E \in \mathbb{R}^{K \times F}$ is not an orthogonal matrix $EE^T \neq \mathbf{I}^K$, since the endmember spectra themselves do not form an orthogonal matrix. This can be easily verified for any dataset with known ground truth endmember spectra.
\end{remark}

The reason for the absence of orthogonality is our desired LMM representation in Eq. \ref{eq:lmm} where the abundances must sum up to one, resulting in a correlated endmember spectra. Consequently, this linear restriction leads to the formation of a $(K-1)$ dimensional simplex \cite{bioucas2013hyperspectral}.

\subsection{Architecture}

\begin{figure}[ht]
    \centering
    \vspace{-1em}
    \includegraphics[width=0.5\linewidth]{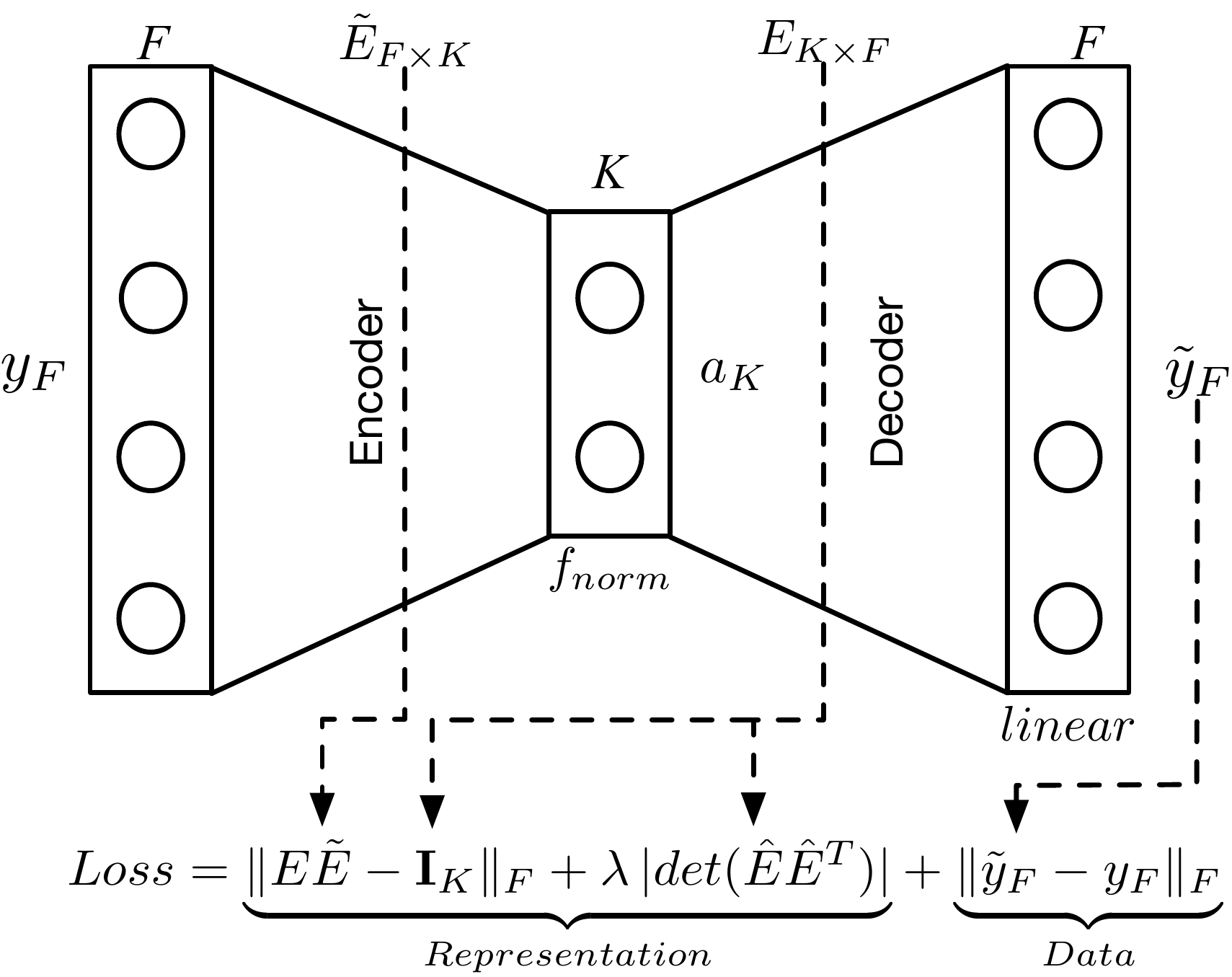}
    \caption{SCA Architecture. The network has only one encoder and one decoder, with two representation driven losses, besides the conventional data reconstruction loss.}
    \vspace{-1em}
    \label{fig:arch}
\end{figure}

Following the previous discussion, SCA architecture consists of only one encoder and one decoder with $K$ and $F$ neurons, respectively. The decoder weights $E$ represent the endmember spectra matrix, whereas the encoder weights $\tilde{E}$ is the pseudo-inverse (conjugate dual) of $E$ once the network minimization problem (\textbf{Section \ref{sec:netmin}}) converges. SCA requires only $2FK$ parameters independent of the sample dimension $N$. \textbf{Fig. \ref{fig:arch}} shows the network architecture and the associated network minimization problem. The encoder has a non-linear activation function (see \textbf{Section \ref{subsec:act}}) while the decoder activation is chosen to be linear satisfying LMM. Since biases account for a mean feature that does not exist in HSI datasets, biases are not used for any layer. 

\begin{remark}
    Note that the encoder and decoder in a sequence is equivalent to the action of the matrices $\tilde{E}$ and $E$, respectively as in Eq. \ref{eq:form}.
\end{remark}

The LMM model suggests that any autoencoder design should have only two layers \textit{iff} the user desires to interpret the spectral data as a weighted combination of endmember spectra. Here, the endmember spectra itself is an invariant of the system that does not change with samples, whereas the abundances are the system variants. The reason for this is the interpretability requirement imposed by the user to decompose the dataset as weighted linear combination of endmembers $E$ corresponding to the fractional abundances $A$. Any deviation from this autoencoder design also implies a different representation of the system itself wherein the invariants can no longer be identified as humanly interpretable endmember spectra. \textit{Note that finding an alternate representation and demonstrating that the representation holds with arbitrary precision is non-trivial.}

\subsection{Network Minimization Problem}\label{sec:netmin}

The network minimization problem consists of three additive terms categorized as data and representation driven loss terms. The first term in the representation loss ensures a bi-orthogonal representation even when a user inadvertently over-prescribes the number of endmembers. Further, the reconstruction and bi-orthogonality loss terms are bounded below (see \textbf{Section \ref{subsec:errbnd}}). Note that with SCA our objective is to find a mathematically sound bi-orthogonal representation $\hat{Y}$ that spans the top rank-K, finite dimensional, vector space of the input data $Y_{F}$. A detailed description of this bi-orthogonal representation and its relationship to Singular Value Decomposition (SVD) can be found in \textbf{Section \ref{subsec:biorth}}.

The second term in the representation loss ensures a minimum volume criterion so as to obtain the smallest $(K-1)$ simplex formed by the endmembers in the feature dimensional space. This minimum volume loss term is further expanded upon in \textbf{Section \ref{subsec:volmin}}. Upon convergence, the endmember matrix ($E$) is extracted as decoder weights where the encoder weights ($\tilde{E}$) form a dual of the endmember matrix such that $E\tilde{E} = \mathbf{I}_{K}$. \textbf{Section \ref{subsec:prop}} presents a brief description of the network properties while characterizing the loss surface of the minimization problem. Finally, SCA does not suffer from over and under-fitting issues due to our choice of a bi-orthogonal representation (see \textbf{Section \ref{subsec:over}}).

\subsection{Choice of Activation Functions}\label{subsec:act}

Our network architecture is low-weights comprising of only one encoder and one decoder. A linear activation is used for the output layer while the hidden layer activation is defined as a normalized rectified linear unit ($f_{norm}$) as follows in \textbf{Eq. \ref{eq:frelu}}:

\begin{equation}
    {f}_{norm,k}(\mathbf{y_F}) = \frac{max(0,y_{k})}{\sum_{k} max(0,y_{k}) + \epsilon} \label{eq:frelu}
\end{equation}

This specialized choice of hidden layer activation function ensures $f_{k} \in [0,1], \forall k = 1,\ldots, K$, to account for pure endmembers. In other words, the map $f_{norm}$ is such that $f: Y_F \in \mathbb{R}^{F} \rightarrow \tilde{Y}_F \in [0,1]^{K}$. Here, $K$ is the number of desired end-members, and $\epsilon \approx 10^{-8}$ (GPU-precision) is a small, non-negative, real number to avoid singularity when $f$ is identically zero for some $\textbf{y}$. We point out that using infinite-dimensional, non-linear activation functions ($g$ = \textit{tanh, softmax}, \etc) will incur approximation errors. One can observe that $g(y)\rightarrow 0$ or $g(y)\rightarrow 1$ as $y \rightarrow \pm \infty$, resulting in higher approximation errors where the dataset consists of samples where at least one abundance is zero. On the other hand, our normalized \textit{relu} function $\mathbf{f}_{norm}$ satisfies $\sum_{k}f_{norm,k} = 1$ at GPU precision that can be altered by changing $\epsilon$. The range of $f_{norm}$ now serves as the pixel abundances. 

Another advantage of using this normalized activation function is that the simplex constraint in Eq. \ref{eq:lmm} is accounted for without introducing an additional loss term and consequently hyper-parameter that must be tuned while training SCA. The loss function corresponding to our network minimization problem therefore contains only three terms to account for reconstruction loss, representation loss and minimum volume criterion with only one hyper-parameter. \textit{Finally, the most important outcome of this choice is that the network weights can now be initialized arbitrarily as opposed to initialization from a feasible set satisfying the simplex constraint.} 

\subsection{Minimum Volume Criterion}\label{subsec:volmin}

As discussed in prior works \cite{bioucas2013hyperspectral}, in order to restrict the solution space $\hat{W}_{K}$ of the endmembers, we require a minimum volume criterion on the $(K-1)$-dimensional simplex formed by $E$ in an $F$-dimensional space. We rely upon mean corrected end-members for volume calculation to avoid conditioning issues and to ensure that the length of the vectors in the determinant calculation are at a similar scale. The mean correction $E_{K,m}$ is:,
\begin{align*}
    E_{K,m} = E - \bar{E}
\end{align*}
A matrix $\hat{E}$ can now be constructed as,
\begin{align*}
    \hat{E} = [\mathbf{1}, E_{1,m},\ldots, E_{K-1,m}] \in \mathbb{R}^{K \times F}
\end{align*}
with the volume given by,
\begin{align*}
    vol = |det(\hat{E}\hat{E}^T)|
\end{align*}
Please note that, this volume calculation is invariant to rotation and translation of the end-member matrix $E$ that later helps us (\textbf{Section \ref{subsec:errbnd}}) to characterize the loss surface of the network minimization problem in Fig. \ref{fig:arch}.

\subsection{Bi-orthogonality and Self-Correction}\label{subsec:biorth}

We now describe a bi-orthogonal representation to approximate $Y_{F}$ with its rank $K$ truncation $\tilde{Y}_{F}$ under LMM constraints. Given a dataset $Y_{F} \in U_{F}$ with $F$ features, let us define a vector space $V_{K} = span\{v_{1},\ldots, v_{k}\}$. Any $V_{K} \in \mathbb{R}^{F \times K}$ can represent $Y_{F}$ approximately as $\tilde{Y}_{F} = Y_{F}V_{K}V_{K}^{T}$ if $V_{K} \subset U_{F}$. Here, $V_{K}$ is the orthogonal vector space spanned by the $K$ right singular vectors of $Y_{F}$ corresponding to the $K$ largest singular values such that $V_{K}^{T}V_{K} = \mathbf{I}_{K}$. However, now there is no room for enforcing the simplex constraint defined in Eq. \ref{eq:lmm} or the minimum volume criterion. 

A bi-orthogonal representation \cite{andrle2007experiments} on the other hand states that, a vector space $H_{K}$ and its bi-orthogonal dual $\tilde{H}_{K}$  can be used to represent $Y_{F}$ approximately as $\tilde{Y}_{F} = Y_{F}\tilde{H}_{K}H_{K}^{T}$ as long as $\tilde{H}_KH_K^{T} = span\{h_{1},\ldots, h_{k}\} \in W_{K}$ is such that $W_{K} \subset Y_{F}$ with $H_{K}^{T}\tilde{H}_{K} = \mathbf{I}_{K}$. This bi-orthogonal representation now gives us flexibility to enforce additional constraints. Under this description, a right bi-orthogonal projection described by $\tilde{H}_{K}H_{K}^{T}$ must have the same span as the top-$K$ right singular vectors $V_{K}$ or $\tilde{H}_{K}H_{K}^{T} = V_{K}V_{K}^{T}$.

Please note $H_{K}$ and $\tilde{H}_{K}$ are non-orthogonal and therefore $H_{K}^{T}H_{K} \neq \mathbf{I}_{K}$ and $\tilde{H}_{K}^{T}\tilde{H}_{K} \neq \mathbf{I}_{K}$. In the following, we show that under a bi-orthogonal representation the approximation error between the input data $Y_{F}$ and its approximation $\tilde{Y}_{F}$ is bounded below by the $(F-K)$ tail energy equipped with a Frobenius norm following Eckart-Young-Mirsky (EYM) \cite{eckart1936approximation} theorem. Upon convergence, the non-orthogonal vectors $H_{*,K}^{T} \approx E$ are the end-members and $\tilde{H}_{*,K} \approx \tilde{E}$ is the dual (psuedo-inverse) of end-member matrix $E$. 

This representation renders our framework a self-correcting property wherein if a user inadvertently over-specifies the number of end-members (K+O) than are present in the data (K), the abundance maps corresponding to this overspecification (O) are identically zero. Our autoencoder formulation is therefore closely related to SVD as discussed later in \textbf{Section \ref{subsec:over}}. A bi-orthogonality loss ($\|E\tilde{E} - \mathbf{I}_{K}\|_{F}$) ensures that the network generated vector space always spans a user-specified (K+O)-dimensional space. 

\subsection{Error Bounds}\label{subsec:errbnd}

EYM theorem \cite{eckart1936approximation} states that the tail energy $\|Y - \tilde{Y}\|_{F}$ of a low rank approximation $\tilde{Y}$ under a Frobenius norm is bounded by:
\begin{align*}
\|Y - \tilde{Y}\|_{F} \geq \|Y - \hat{Y}\|_{F}
\end{align*}
where, $\hat{Y}$ is the top rank-K approximation of $Y$ corresponding to the top-K singular values of $Y$. Substituting $\tilde{Y} = Y\tilde{H}_{K}H_{K}^{T}$ we obtain lower bounds on our low-rank bi-orthogonal representation as,

\begin{equation}
    \|Y - Y\tilde{H}_{K}H_{K}^{T}\|_{F} \geq \|Y - \hat{Y}\|_{F}.
    \label{eqn:recons}
\end{equation}

Additionally, we also know that the bi-orthogonality loss and minimum volume criterion are bounded below: 

\begin{equation}
    \|H_{K}^{T}\tilde{H}_{K} - \mathbf{I}_{K}\|_{F} \geq 0
    \label{eqn:biorth}
\end{equation}

From Eq. \ref{eqn:recons} and \ref{eqn:biorth} we have,
\begin{equation}
    \|Y - Y\tilde{H}_{K}H_{K}^{T}\|_{F} + \|\tilde{H}_{K}H_{K}^{T} - \mathbf{I}_{K}\|_{F} \geq \|Y - \hat{Y}\|_{F}
\end{equation}

Here, equality is achieved when $span\{\tilde{H}_{*,K}H_{*,K}^{T}\} = span\{\hat{Y}\} = span\{v_{1},\ldots,v_{k}\}$ where $v_{i}$s are the top-K right singular vectors of $Y$. The data driven loss in the network minimization problem must therefore converge to this tail energy for us to extract $\tilde{E} = \tilde{H}_{*,K}$ and $E = H_{*,K}^{T}$. With the current network architecture if one chooses a $linear$ or $relu$ activation for the hidden layer, this tail energy can be computationally verified upon convergence.

However, the unmixing problem also requires that the abundances or the encoder outputs ($A$) satisfy a simplex criterion such that $\sum_{k}a_{k}$ is identically 1 where the entries $a_{k} \geq 0$. Our choice of hidden layer activation function ensures this criterion is satisfied as described before in Subsection \ref{subsec:act}. We would like to point out that in the absence of the simplex and minimum volume criteria, the vectors spaces $\tilde{H}_{*,K}$ and $H_{*,K}$ are not unique although the space spanned by $\tilde{H}_{*,K}H_{*,K}^{T}\in W_{K}$ is unique and is a subset of the space spanned by $Y_{K}$ or $W_{K} \subset Y_{K}$. The constraints now result in a smaller subset $\hat{W}_{K} \subset W_{K}$ from where an approximate solution can now be extracted. In this respect, our choice of hidden layer activation is closely related to projected gradient descent (PGD) method \cite{nocedal2006numerical} for optimization problems where a projection of the network-weights update on a feasible space satisfying the constraints is computed. 

\subsection{Network Properties}\label{subsec:prop}
Given the ground truth abundances and end-members, our interpretable network-weights can be trivially initialized to find that: \textbf{1)} the network weights do not alter upon training and \textbf{2)} the data driven loss term matches the EYM tail energy bound at GPU precision. \textit{This serves as a verification step for our proposed formulation where the true solution is a global minimum of the network minimization problem.} We now discuss a few additional properties of our network minimization problem:
\begin{enumerate}[leftmargin=*]
    \item Our definition of the minimum volume criterion is invariant to rotation and translation due to a mean correction of end-members prior to volume calculation. Since the end-members form a convex simplex in an $F$ dimensional space, the mean of the end-members always lies inside the simplex and therefore a volume calculation with the mean of end-members as the frame of reference renders this invariance.
    \item The number of achievable global minima are $K!$. Considering a $K$ end-member problem where $E$ and $A$ are the end-member and abundance matrices with $\{e_{1},\ldots,e_{K}\}$ and $\{a_{1},\ldots,a_{K}\}$ vectors, we know that any one of the $k = 1,\ldots K!$ permutations of $e_{k}$s and $a_{k}$s satisfies $Y_F = A_{k}E_{k}$. Therefore, in the constraint satisfying subspace $\hat{W}_{K}$ the loss surface has $K!$ global minima of equal energies. 
    \item The equal energy global minima are bounded by the rank-K tail energy as discussed before and can be computationally verified upon convergence.
\end{enumerate}

\begin{remark}
A trivial check to see if SCA network formulation is correct is to initialize the decoder weights by ground-truth endmember matrix $E$ and encoder weights by the right pseudo-inverse of $E$. This provides us a verification step where the identity $0=0$ corresponding to $Y_F-AE = 0$ is satisfied at GPU precision.
\end{remark}


\subsection{Low-weights and Over-fitting}\label{subsec:over}

Under our problem formulation the number of trainable network parameters is known \textit{a priori} as $2FK$ where $F$ is the number of features in the input data $Y_{F}$ and $K$ are the number of desired or prescribed end-members. Since the network architecture is guided by a bi-orthogonal representation similar to SVD, we do not face over- or under-fitting issues upon convergence even when the user prescribes an over-specified number of end-members ($K+O$). \textit{As a consequence, our low weight and interpretable network architecture does not require fail-safe measures such as dropout, batch-normalization, or kernel regularization as additional avenues to achieve higher accuracy.}

Let us consider a noiseless HSI data matrix $Y_F$ with $F$ features and $K$ endmembers (known \textit{a priori}). A Singular Value Decomposition of $Y_F$ then indicates $K$ non-zero singular values and $F-K$ zero singular values. Consequently, $F-K$ singular vectors (left and right) belong to the null space and therefore the vectors themselves can be arbitrary since the singular value itself is zero. The left ($U_{K}) $ and right ($V_{K})$ singular matrices formed by the orthogonal vectors corresponding to the non-zero $K$ singular values can then be used to represent the data matrix $Y_F$ exactly as $Y_F = U_{K}\Sigma_{K}V_{K}^{T} = Y_FV_{K}V_{K}^{T}$. One can easily check that $V_{K}^{T}V_{K} = \mathbf{I}_{K}$. Any rank $K+O$ approximation of $Y_F$ is then $Y_F = U_{K+O}\Sigma_{K+O}V_{K+O}^{T} = Y_FV_{K+O}V_{K+O}^{T}$ since the $O$ singular values are all zero. 

However, we are already aware that the $K$ right singular vectors of $Y_F$ do not form the endmember matrix since the end-member spectra themselves are not necessarily orthogonal to each other. Here, bi-orthogonality renders  flexibility by relaxing the orthogonality restriction on the matrix factors by constructing $Y_F = Y_F\tilde{E}_{K}E_{K}$ such that $E_{K}\tilde{E}_{K} = \mathbf{I}_{K}$ similar to SVD. Substituting $Y_F = AE$ on the right hand side now results in $Y_F = AE\tilde{E}_{K}E_{K} = A_{K}E_{K}$. Similar to the over-specified $K+O$ in the previous paragraph, $Y_F = AE\tilde{E}_{K+O}E_{K+O} = A_{K+O}E_{K+O}$ where abundances $A_{O}$ corresponding to the overspecification are identically zero (compare to singular values above) with arbitrary, null space, endmember spectra $E_{O}$ and it's bi-orthogonal dual $\tilde{E}_{O}$. As before, since the abundances corresponding to the over-specified endmember spectra ($O$) do not contribute the representation still remains exact. For the under-specified endmembers $K-U$, SCA returns a low rank approximation of $Y_F$ that still abides by all the LMM constraints. 
The reader is referred to numerical results in \textbf{Section \ref{sec:over}} for the over-specified endmember case using Samson dataset.


\subsection{Storage Complexity Analysis}

Conventional matrix factorization based approaches for LMM involve decomposing the entire data matrix $Y_F \in \mathbb{R}^{N \times F}$ into the abundance $A \in \mathbb{R}^{N \times K}$ and endmember $E \in \mathbb{R}^{K \times F}$ matrices leading to a memory cost proportional to $\mathcal{O}(NK+KF)$. An explicit advantage of a neural approach is that the abundance matrix $A$ need not be present in the main memory at all. Compared to other neural endmember extraction approaches \cite{ozkan2018endnet,su2019daen,khajehrayeni2020hyperspectral}, SCA has an exact memory requirement of $\mathbf{2FK}$ parameters. As mentioned before, the LMM system is not reducible beyond this parameter requirement without adversely affecting the error metrics or interpretability. SCA is also fully interpretable since the decoder weights form the endmember matrix with the encoder wights storing the right pseudoinverse of the endmember matrix.

\section{Results} \label{sec:results}

Here we describe our training setup and outputs for three HSI datasets and profiling against the state of art methods in terms of error metrics. 
Finally, we profile the runtime requirements, and convergence of SCA.

\subsection{Setup and Training}

All experiments were done on a setup with Nvidia 2060 RTX Super 8GB GPU, Intel Core i7-9700F 3.0GHz 8-core CPU and 16GB DDR4 memory. We use the Keras \cite{chollet2015} library running on a Tensorflow 1.15 backend with Python 3.7 to train the networks in this paper. For optimization, we use AdaMax \cite{kingma2014adam} with parameters (\textit{lr}= 0.0001) and $1000$ steps per epoch. A common trend in neural approaches is to pre-initialize the network with layer-wise training \cite{bengio2007greedy}. The strength of SCA is that all network weights are initialized by drawing from a random uniform distribution every run, yet the network converges to the true solution with high precision.

\subsection{Training and Validation Split} \label{sec:trval}

An issue with training and validation split in matrix decomposition problems is that the error norm cannot be bounded in a deterministic manner or computationally verified. For example, a Singular Value Eecomposition of a given data matrix $Y_F$ differs from SVD on a truncated dataset $\hat{Y}$ in it's singular triplets (singular values and vectors). Ensuring these triplets do not change over an arbitrary split is a non-trivial computational task.
\begin{remark}
For dataset $Y_F$, an arbitrary training/validation split results in a varying dataset $\hat{Y}$ wherein the norm $\|Y_{pred}-\hat{Y}\|_{F}$ changes according to the split. Since the desired features are unknown \textit{a priori}, a consistent truncated dataset $\hat{Y}_{c}$ that spans the same space as the full data $Y_F$ cannot be obtained using an arbitrary split.
\end{remark}
This results in a large variance in extracted features over multiple training/validation splits since the span of $\hat{Y}$ itself is changing with each split. Furthermore, the minimum volume criterion becomes adversely sensitive to this changing span of the dataset. Our errors on the other hand are bounded since we do not perform a training-validation split so as to bound the error as $\|Y_{pred}-Y\|_{F}$, where $Y_F$ is the static dataset (frame of reference). Our neural architecture consistently arrives at a low-error approximation, over multiple runs, by passing through entire dataset batch-wise.

\subsection{Pre and Post-Processing}

The data $Y_F$ is pre-processed to scale of $[0,1]^{N \times F}$:
\begin{align*}
    Y_{s} = \frac{Y_F - min(Y_F)}{max(Y_F)-min(Y_F)}.
\end{align*}
This ensures that the reconstruction and bi-orthogonality losses $\|Y_F - Y_F\tilde{H}_{K}H_{K}^{T}\|_{F}$ and $\|\tilde{H}_{K}H_{K}^{T}-\mathbf{I}_{K}\|_{F}$, respectively are at a similar scale to avoid conditioning issues and consequently precision errors. Upon convergence we obtain a scaled endmember matrix and its dual as $H_{*,K}^{T} = E_{s}$ and $\tilde{H}_{*,K} = \tilde{E}_{s}$, respectively. The scaled endmember matrix $E_{s}$ can now be un-scaled to arrive at the true end-members as follows:
\begin{align*}
    E = E_{s}\left(max(Y_F)-min(Y_F)\right) + min(Y_F)
\end{align*}

Note that, compared to other end-member extraction formulations \cite{ozkan2018endnet,su2019daen,khajehrayeni2020hyperspectral} our pre- and post-processing steps only comprise of linear scaling without altering the abundances at a trivial computational cost. Specifically, Endnet extracts abundances that do not sum up to 1 and consequently an additional algorithm \cite{heylen2014nonlinear} is required to post-process the abundances $A$. Also note that for Endnet post-processing the abundances $A$ alone, to ensure the simplex criterion is satisfied, does not guarantee that the post-processed $A$ now satisfies $Y_F = AE$ from Eq. \ref{eq:lmm}. 

\subsection{Metrics}

To evaluate unmixing performance against the ground truth, we utilize two metrics: Spectral Angle Distance (SAD) and Root Mean Square Error (RMSE). Smaller values indicate better performance for both metrics.

\begin{minipage}{\linewidth}
\begin{align*}
    \textrm{SAD}(x, \hat{x}) = \cos^{-1} \left ( \frac{x.\hat{x}}{\|x\|_2\|\hat{x}\|_2} \right ),     \textrm{RMSE}(x, \hat{x}) = \sqrt{\frac{1}{N}\|x-\hat{x}\|_2^2}
\end{align*}
\end{minipage}

\subsection{Overall Analysis}

\begin{table}[ht]
    \centering
    \begin{tabular}{cc|cccc}
    \toprule
        \multirow{2}{*}{Data} & \multirow{2}{*}{Members} & \multicolumn{2}{c}{EndNet $(10^{-2})$} & \multicolumn{2}{c}{Ours} \\
        & & RMSE(A) & SAD(E) & RMSE(A) $(\mathbf{10^{-5}})$ & SAD(E) $(\mathbf{10^{-4}})$ \\ \midrule
        \multirow{3}{*}{Samson} & Mem1 & 5.72$\pm$0.0 & 1.29$\pm$0.1 & \textbf{1.69$\pm$0.1} & \textbf{2.06$\pm$0.0} \\
                                & Mem2 & 3.84$\pm$0.1 & 4.69$\pm$0.1 & \textbf{1.67$\pm$0.1} & \textbf{1.13$\pm$0.0} \\
                                & Mem3 & 2.11$\pm$0.0 & 2.95$\pm$0.3 & \textbf{0.21$\pm$0.0} & \textbf{1.68$\pm$0.0} \\ \midrule
                                & Avg. & 3.88$\pm$0.0 & 2.98$\pm$0.2 & \textbf{1.18$\pm$0.1} & \textbf{1.69$\pm$0.0} \\ 
        \midrule\midrule
        \multirow{4}{*}{Jasper} & Mem1 & 8.24$\pm$0.4 & 4.99$\pm$0.4 & \textbf{1.91$\pm$0.1} & \textbf{2.20$\pm$0.0} \\
                                & Mem2 & 6.17$\pm$0.3 & 4.23$\pm$0.9 & \textbf{3.81$\pm$0.1} & \textbf{4.20$\pm$0.0} \\
                                & Mem3 & 8.98$\pm$0.2 & 4.47$\pm$0.3 & \textbf{3.18$\pm$0.2} & \textbf{0.00$\pm$0.0} \\
                                & Mem4 & 8.55$\pm$0.1 & 1.96$\pm$0.2 & \textbf{4.48$\pm$0.1} & \textbf{4.14$\pm$0.0} \\ \midrule
                                & Avg. & 7.96$\pm$0.3 & 3.91$\pm$0.5 & \textbf{3.34$\pm$0.1} & \textbf{2.63$\pm$0.0} \\ 
        \midrule\midrule
        \multirow{4}{*}{Urban} & Mem1 & 10.41$\pm$0.2 & 6.88$\pm$0.2 & \textbf{2.63$\pm$0.1} & \textbf{2.43$\pm$0.0} \\
                               & Mem2 & 12.24$\pm$0.3 & 3.92$\pm$0.3 & \textbf{0.96$\pm$0.0} & \textbf{0.75$\pm$0.0} \\
                               & Mem3 & 8.35$\pm$0.3 & 3.53$\pm$0.1 & \textbf{1.26$\pm$0.1} & \textbf{2.57$\pm$0.0} \\
                               & Mem4 & 5.92$\pm$0.1 & 3.35$\pm$0.5 & \textbf{1.38$\pm$0.1} & \textbf{0.00$\pm$0.0} \\ \midrule
                               & Avg. & 9.23$\pm$0.2 & 4.42$\pm$0.3 & \textbf{1.56$\pm$0.1} & \textbf{1.43$\pm$0.0} \\ 
        \bottomrule
    \end{tabular}
    \caption{Error metrics of SCA \vs Endnet (state of the art). While both metrics for Endnet stagnates at $10^{-2}$, we achieve significantly lower errors at scales $10^{-5}$ and$10^{-4}$ for RMSE and SAD metrics, respectively. Note that Endnet \cite{ozkan2018endnet} does not compute the abundances by their network, but as a post processing step using a different algorithm. SCA on the other hand jointly computes both endmembers and abundances through a single minimization problem abiding LMM constraints.}
    \label{tab:comp-rmse}
    \vspace{-2mm}
\end{table}
\textbf{Table \ref{tab:comp-rmse}} shows the error values for our proposed SCA against the state of the art Endnet \cite{ozkan2018endnet} results. Note that the current best performing models in literature are only able to achieve RMSE(A) and SAD(E) error values at a scale of $10^{-2}$ (\textbf{Table \ref{tab:comp-soa}}). SCA outperforms all of the prior works by two order of magnitude with RMSE(A) and SAD(E) error values at a scale of $10^{-5}$ and $10^{-4}$, respectively. For all the three dataset, the volume penalization parameter $\lambda$ was set to $0.001$. \textbf{Table \ref{tab:our-metric}} shows all the measurable error quantities for SCA.

\begin{table}[ht]
    \centering
    \begin{tabular}{c|cccc}
    \toprule
         & RMSE(Y) (10$^{-4}$) & RMSE(E) (10$^{-5}$) & SAD(E) (10$^{-4}$) & RMSE(A) (10$^{-5}$) \\ \midrule
        Samson & 0.29$\pm$0.0 & 0.48$\pm$0.0 & 1.69$\pm$0.0 & 1.18$\pm$0.1 \\
        Jasper & 1.82$\pm$0.1  & 2.24$\pm$0.1  & 2.62$\pm$0.0  & 3.34$\pm$0.1  \\
        Urban & 0.04$\pm$0.0 & 0.13$\pm$0.0 & 1.43$\pm$0.0 & 1.56$\pm$0.1 \\ \bottomrule
    \end{tabular}
    \caption{SCA error metrics for HSI datasets}
    \label{tab:our-metric}
    \vspace{-2mm}
\end{table}

\subsection{Case: Samson}

\begin{figure}[ht]
    \centering
    \vspace{-1em}
    \begin{subfigure}{\linewidth}
        \centering
        \includegraphics[width=.75\linewidth]{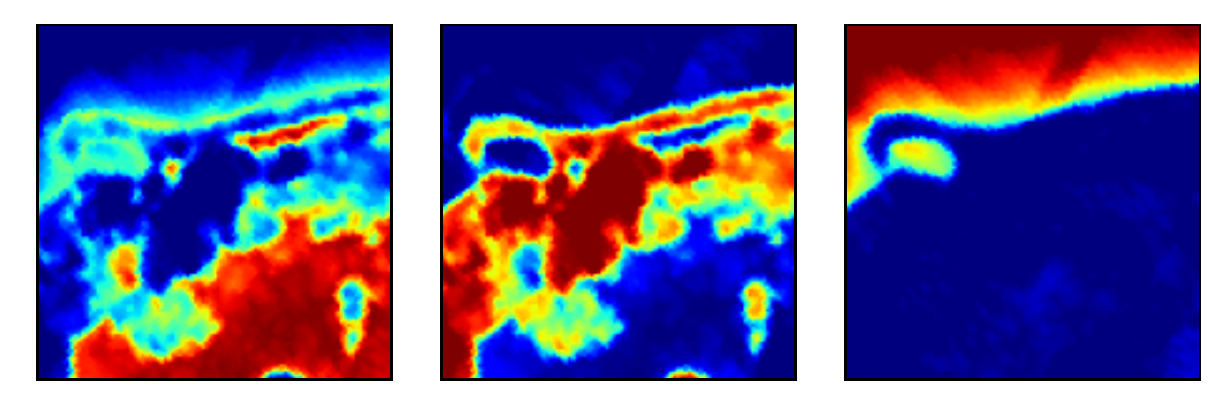}  
    \end{subfigure}
    \begin{subfigure}{\linewidth}
        \centering
        \includegraphics[width=.75\linewidth]{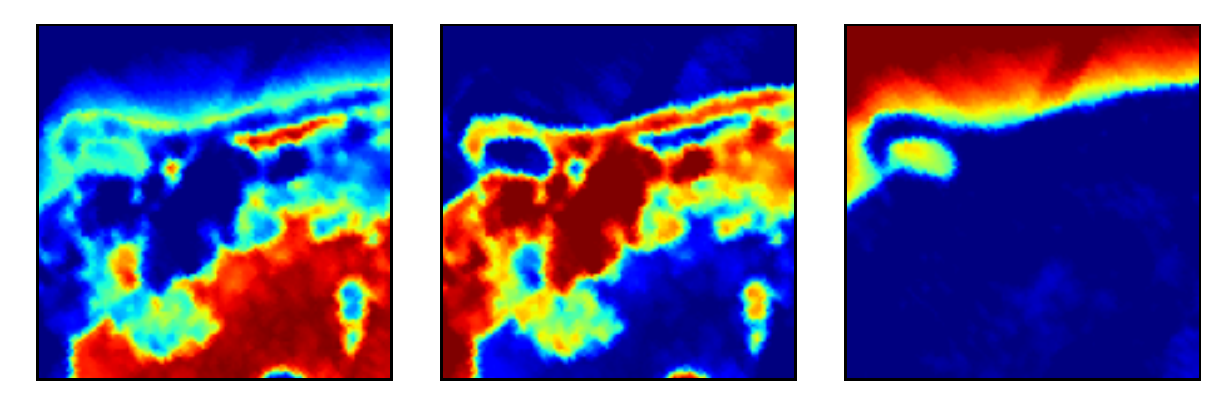}  
    \end{subfigure}
    \begin{subfigure}{\linewidth}
        \centering
        \includegraphics[width=.75\linewidth]{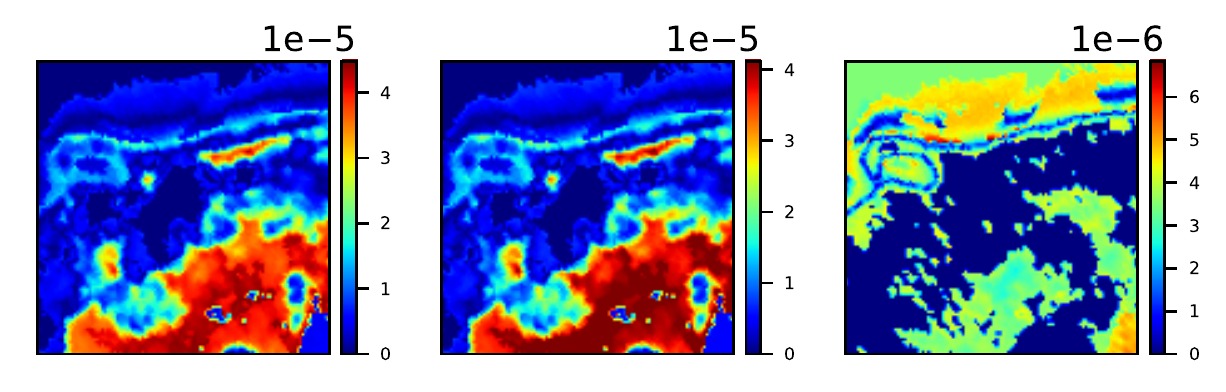}  
    \end{subfigure}
    \vspace{-0.5em}
    \caption{SCA Extracted (top), ground-truth (middle), and absolute difference (bottom) of abundances for Samson dataset. Note that the local errors in the absolute difference maps are at a scale of $10^{-5}$ wherein others report only global errors (RMSE) in each of the maps.}
    \vspace{-1em}
    \label{fig:samson_abun}
\end{figure}

\begin{figure}[ht]
    \centering
    \vspace{-0.5em}
    \includegraphics[width=0.975\linewidth]{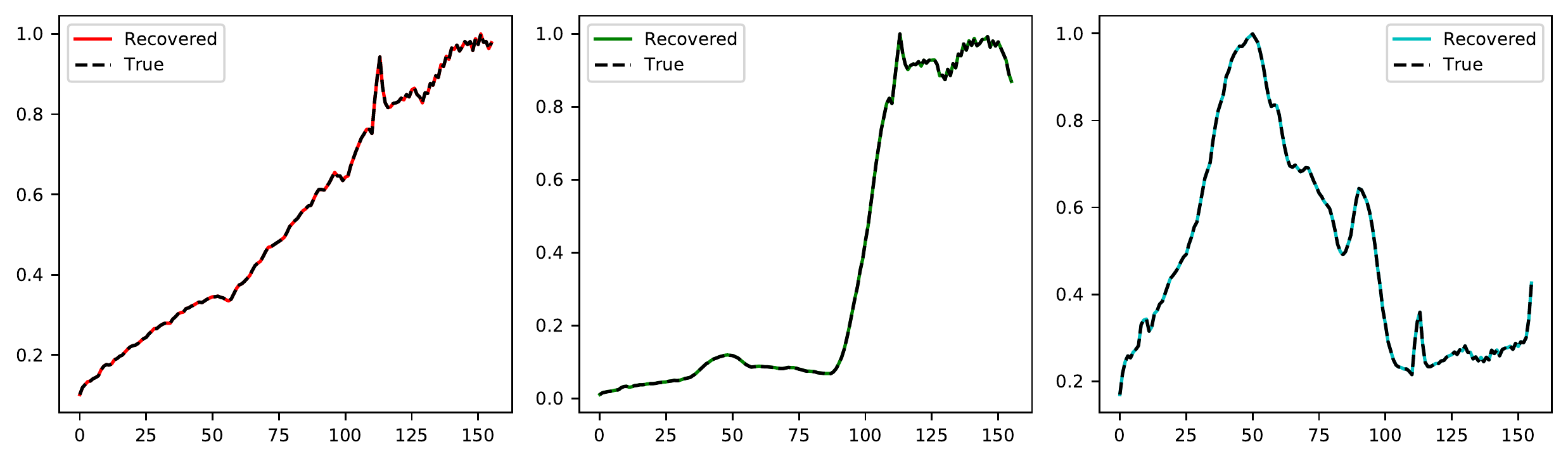}
    \vspace{-0.5em}
    \caption{Extracted (solid line) and ground-truth (dashed-line) endmember spectra for Samson dataset. The two spectra overlap indicating an excellent agreement.}
    \vspace{-1em}
    \label{fig:samson_spec}
\end{figure}

The Samson dataset contains $95 \times 95$ pixels and $156$ channels. 
There are three endmembers: \textbf{Soil}, \textbf{Tree}, and \textbf{Water}. The SCA extracted, ground-truth and absolute difference abundance maps are shown in \textbf{Fig. \ref{fig:samson_abun}} top, middle, and bottom, respectively. The absolute difference abundance maps are at a scale of $10^{-5}$ demonstrating excellent agreement of our solution with the ground-truth. \textbf{Fig. \ref{fig:samson_spec}} shows the extracted endmember spectra (solid lines) overlap with the ground-truth endmember spectra (dashed lines) at error scale $10^{-4}$. 

\subsection{Case: Jasper}

\begin{figure}[ht]
    \centering
    \vspace{-1em}
    \begin{subfigure}{\linewidth}
        \centering
        \includegraphics[width=.95\linewidth]{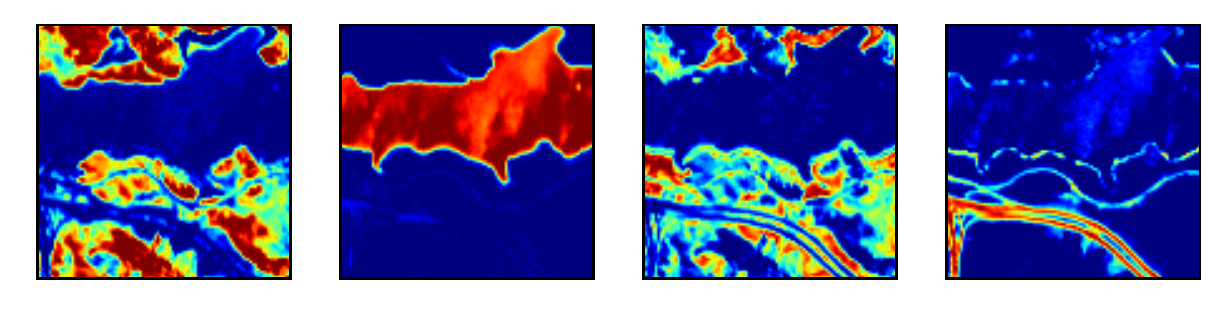}  
    \end{subfigure}
    \begin{subfigure}{\linewidth}
        \centering
        \includegraphics[width=.95\linewidth]{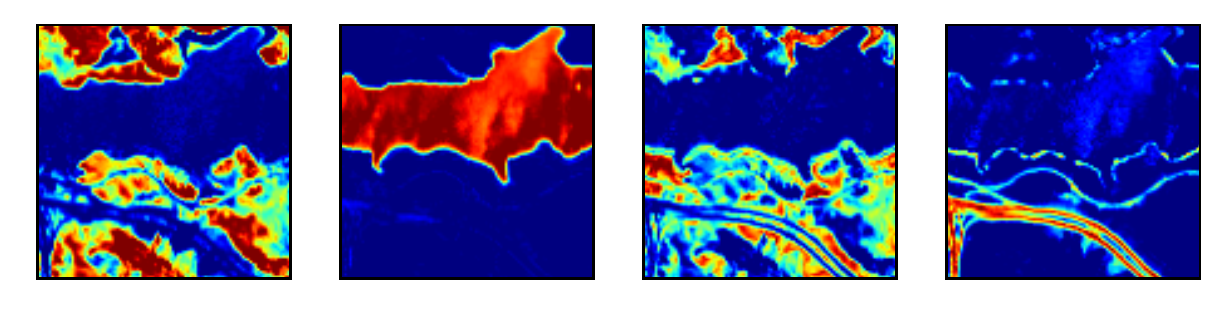}  
    \end{subfigure}
    \begin{subfigure}{\linewidth}
        \centering
        \includegraphics[width=.95\linewidth]{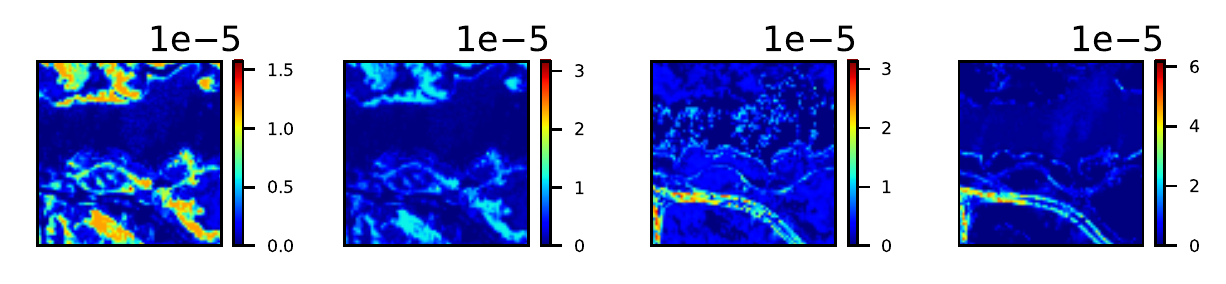}  
    \end{subfigure}
    \vspace{-0.5em}
    \caption{SCA Extracted (top), ground-truth (middle), and absolute difference (bottom) of abundances for Jasper dataset. Note that the local errors in the absolute difference maps are at a scale of $10^{-5}$ wherein others report only global errors (RMSE) in each of the maps.}
    \vspace{-1em}
    \label{fig:jasper_abun}
\end{figure}

\begin{figure}[ht]
    \centering
    \vspace{-1em}
    \includegraphics[width=0.975\linewidth]{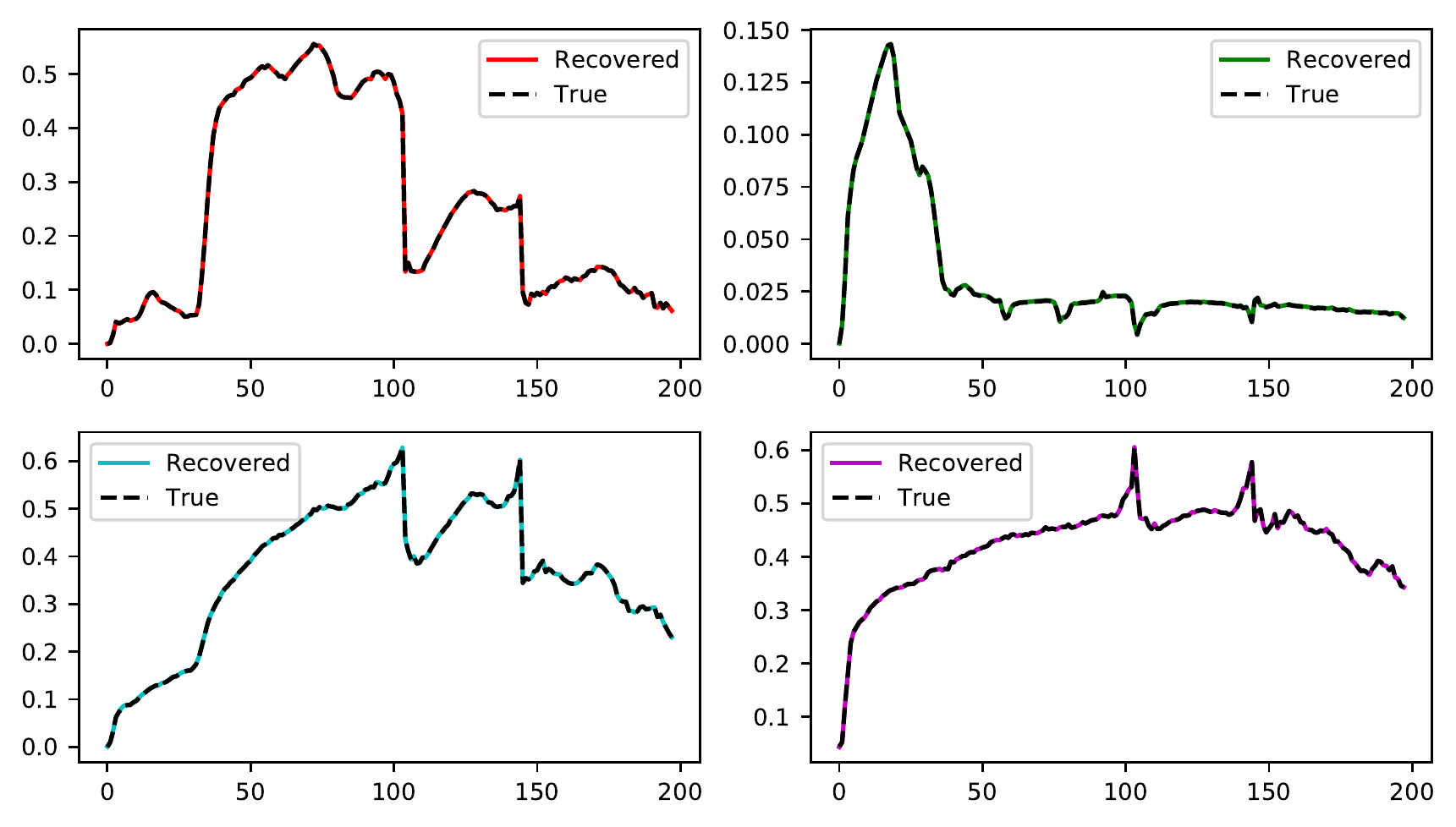}
    \vspace{-0.5em}
    \caption{Extracted (solid line) and ground-truth (dashed-line) endmember spectra for Jasper dataset. Note that the two spectra overlap indicating an excellent agreement.}
    \vspace{-1em}
    \label{fig:jasper_spec}
\end{figure}

The Jasper dataset 
contains $100 \times 100$ pixels and $224$ channels. There are four endmembers: \textbf{Tree}, \textbf{Water}, \textbf{Dirt} and \textbf{Road}. \textbf{Figs. \ref{fig:jasper_abun}} and \textbf{\ref{fig:jasper_spec}} show the abundances and endmember spectra for the Jasper dataset. Again note the excellent agreement with ground truth for both figures with RMSE at $10^{-5}$ and SAD at $10^{-4}$ respectively.

\subsection{Case: Urban}
\begin{figure}[ht]
    \centering
    \begin{subfigure}{\linewidth}
        \centering
        \includegraphics[width=.95\linewidth]{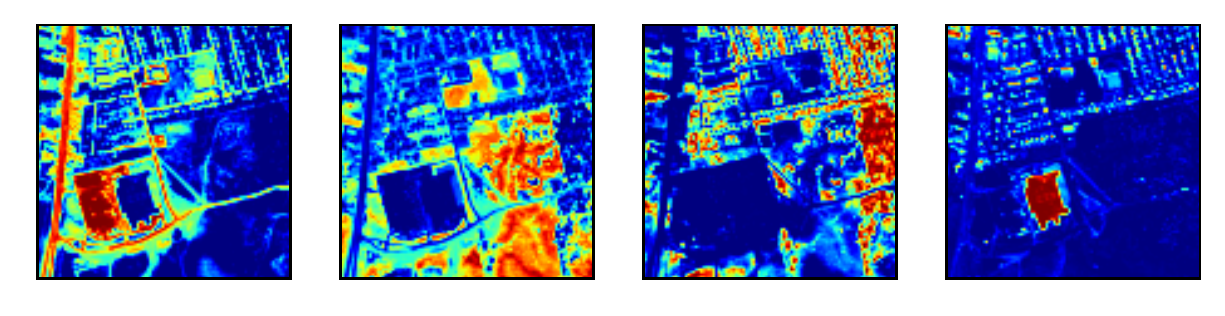}  
    \end{subfigure}
    \begin{subfigure}{\linewidth}
        \centering
        \includegraphics[width=.95\linewidth]{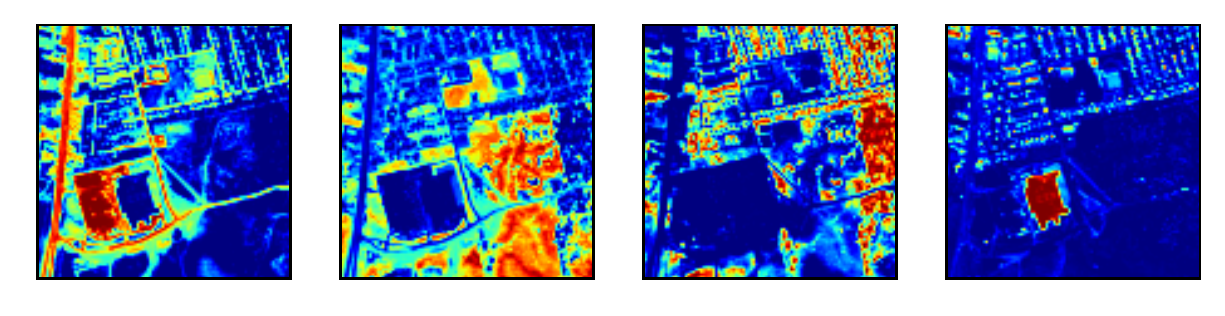}  
    \end{subfigure}
    \begin{subfigure}{\linewidth}
        \centering
        \includegraphics[width=.95\linewidth]{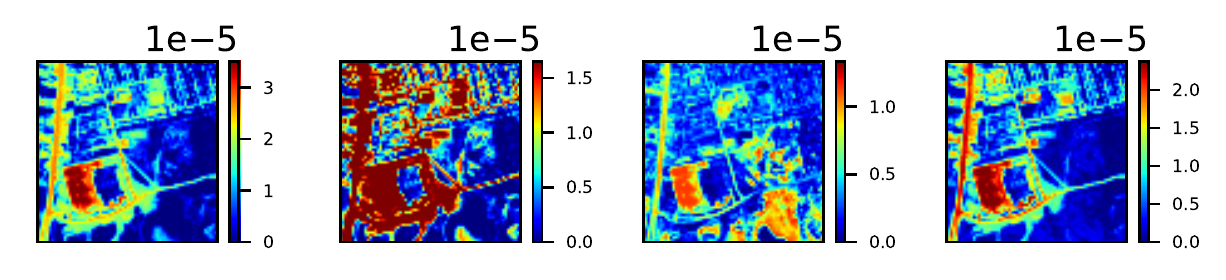}  
    \end{subfigure}
    \vspace{-0.5em}
    \caption{SCA Extracted (top), ground-truth (middle), and absolute difference (bottom) of abundances for Urban dataset. Note that the local errors in the absolute difference maps are at a scale of $10^{-5}$ wherein others report only global errors (RMSE) in each of the maps.}
    \label{fig:urban_abun}
\end{figure}

\begin{figure}[ht]
    \centering
    \includegraphics[width=0.975\linewidth]{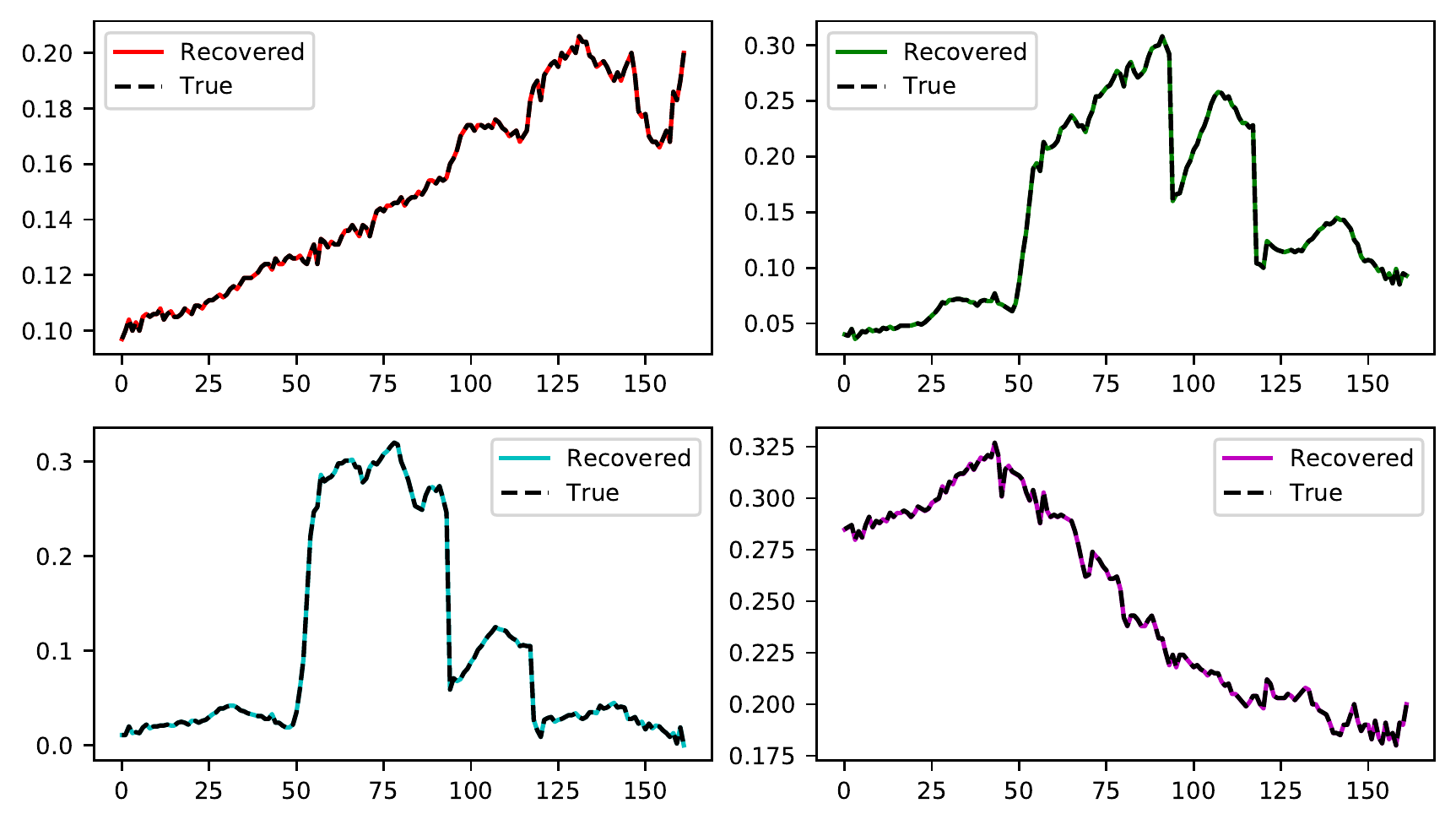}
    \vspace{-0.5em}
    \caption{Extracted (solid line) and ground-truth (dashed-line) endmember spectra for Urban dataset. Note that the two spectra overlap indicating an excellent agreement.}
    \vspace{-1em}
    \label{fig:urban_spec}
\end{figure}

The Urban dataset contains $307 \times 307$ pixels and $162$ channels. 
There are four endmembers: \textbf{Asphalt}, \textbf{Grass}, \textbf{Tree}, and \textbf{Roof}. The extracted, ground-truth and absolute difference abundance maps are shown in \textbf{Fig. \ref{fig:urban_abun}} top, middle, and bottom, respectively. \textbf{Fig. \ref{fig:urban_spec}} shows the extracted endmember spectra (solid lines) overlap with the ground-truth endmember spectra (dashed lines). Figures have RMSE at $10^{-5}$ and SAD at $10^{-4}$ respectively.

\subsection{Self-Correcting Property} \label{sec:over}

\begin{figure}[hb]
    \centering
    \vspace{-1em}
    \begin{subfigure}{0.32\linewidth}
        \centering
        \includegraphics[width=\linewidth]{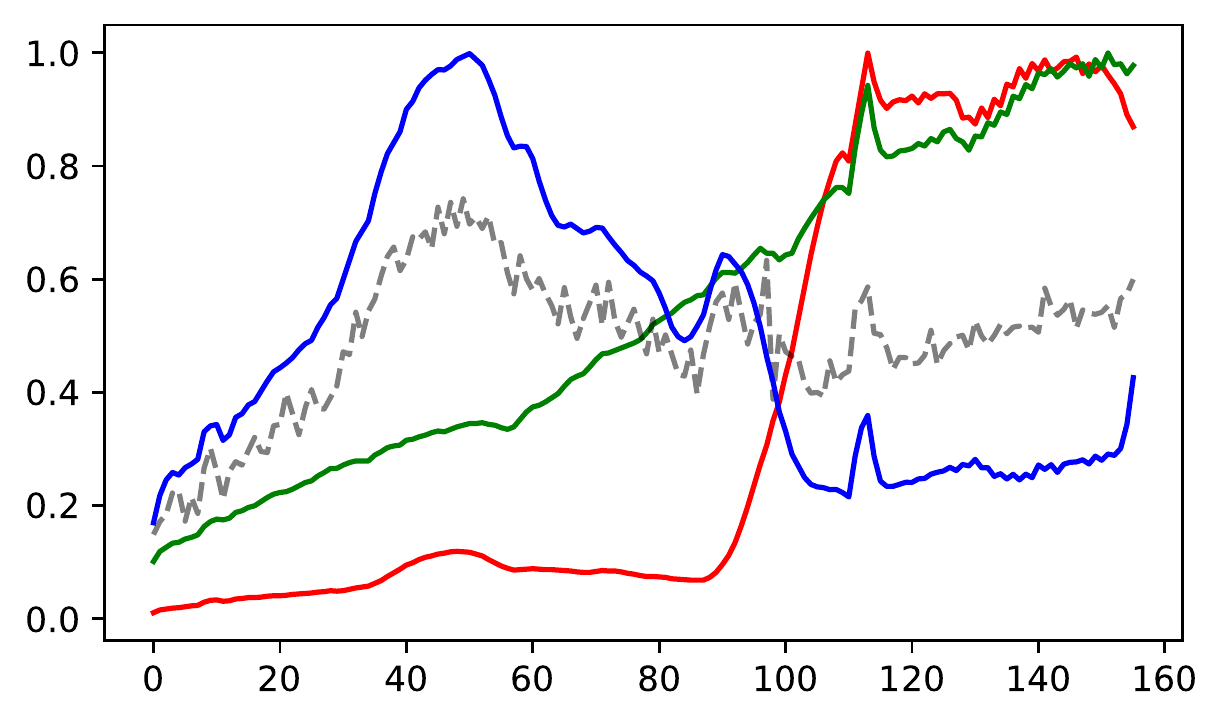}
        \caption{$K+O=4$}
    \end{subfigure}
    \begin{subfigure}{0.32\linewidth}
        \centering
        \includegraphics[width=\linewidth]{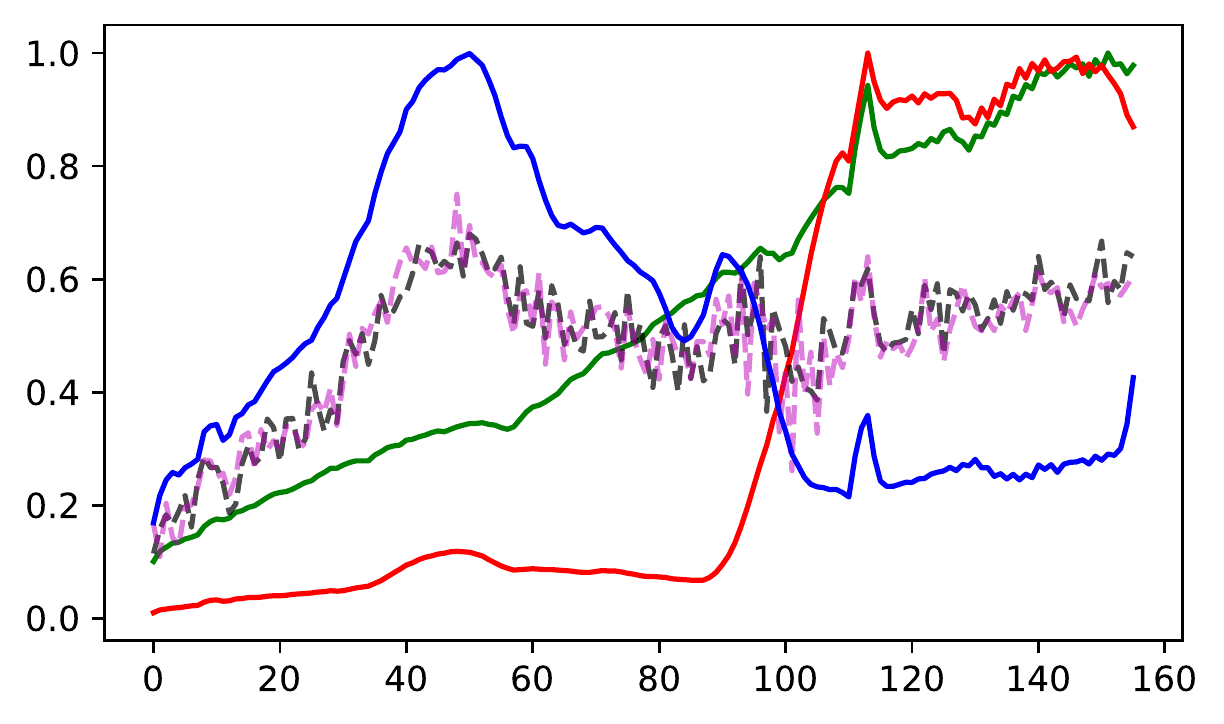}
        \caption{$K+O=5$}
    \end{subfigure}
    \begin{subfigure}{0.32\linewidth}
        \centering
        \includegraphics[width=\linewidth]{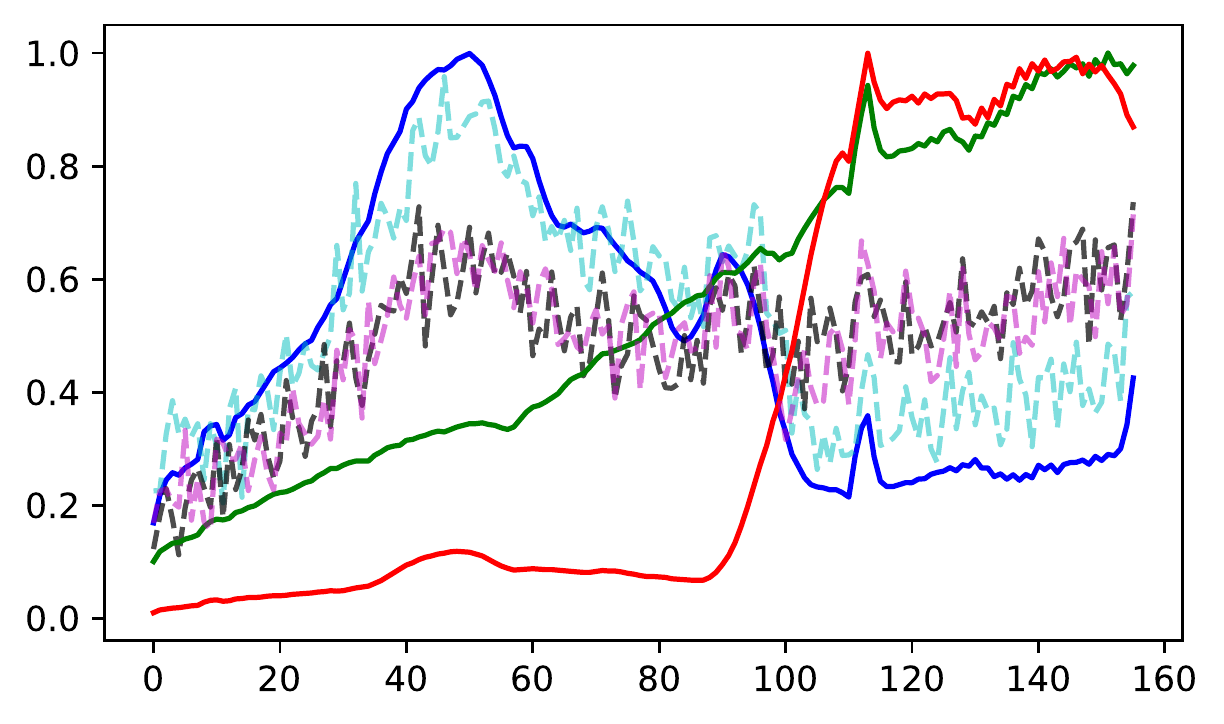}
        \caption{$K+O=6$}
    \end{subfigure}
    \vspace{-0.5em}
    \caption{Extracted endmembers for Samson dataset with over-specified $K$. SCA always retrieves the three true Samson end-members and $K+O-3$ random spectra.}
    \vspace{-0.5em}
    \label{fig:samson_spec_correct}
\end{figure}

In a practical exploratory scenario, the user might not know the correct number of endmembers \textit{a priori}. In such a case, we suggest over-specifying the number of endmembers since SCA's self-correcting property, as discussed in \textbf{Section \ref{subsec:biorth}}, will still extract the correct endmembers and abundances. Additionally, SCA will generate identically zero (GPU precision) abundances for the over-specified endmembers indicating the number of redundant endmembers. In the following, we consider three separate numerical experiments using Samson dataset where the number of ground-truth endmembers is known to be $K=3$. Let us consider three over-specified endmembers as 1) $K+1$, 2) $K+2$, and 3) $K+3$. 

\begin{figure}[ht]
    \centering
    \begin{subfigure}{\linewidth}
        \centering
        \includegraphics[width=0.8\linewidth]{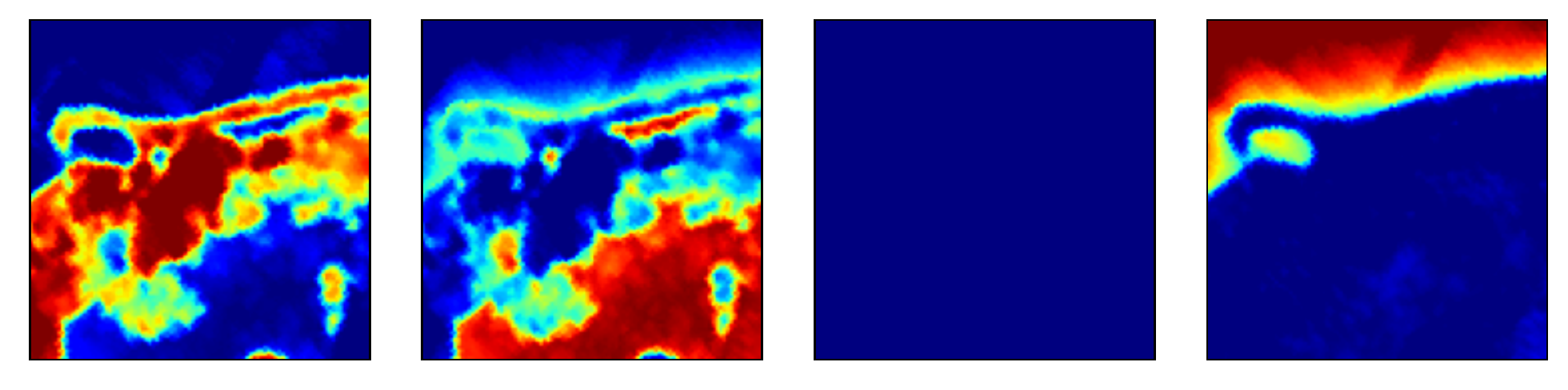}
    \end{subfigure}
    \begin{subfigure}{\linewidth}
        \centering
        \includegraphics[width=0.8\linewidth]{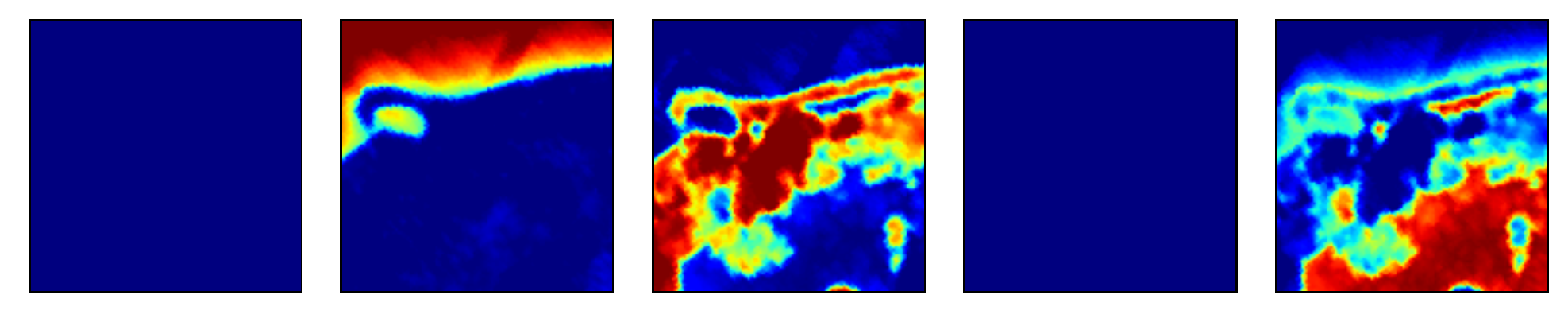}  
    \end{subfigure}
    \begin{subfigure}{\linewidth}
        \centering
        \includegraphics[width=0.8\linewidth]{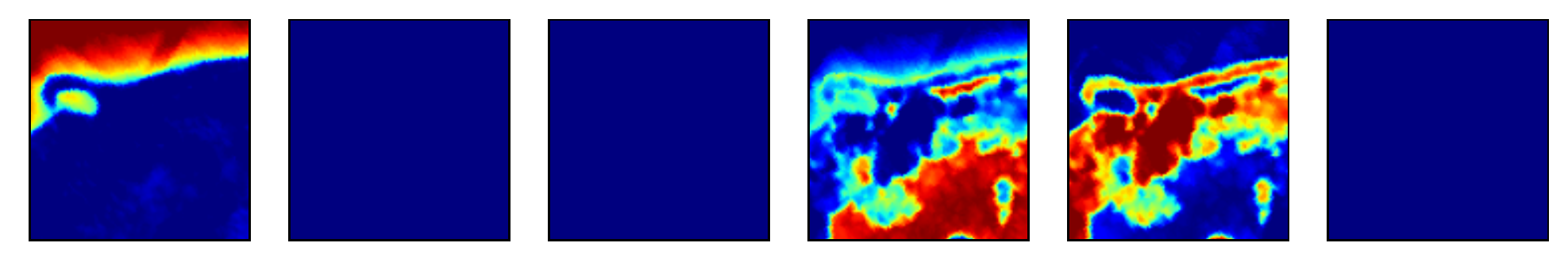}  
    \end{subfigure}
    \caption{Abundance maps for over-specified endmembers $K=4$ (top), $K=5$ (middle), and $K=6$ (bottom).}
    \label{fig:samson_abun_correct}
    \vspace{-5mm}
\end{figure}

\textbf{Fig. \ref{fig:samson_spec_correct}} (left to right) shows the extracted endmember spectra for all three cases. \textbf{Fig. \ref{fig:samson_abun_correct}} shows the corresponding abundance (top to bottom) for these three cases. Note that, the bi-orthogonality loss ensures that the endmember spectra are not identically zero and span a rank $K+O$ space, therefore the additional spurious spectra show up in \textbf{Fig. \ref{fig:samson_spec_correct}}. However, our formulation also ensures that the abundances corresponding to the redundant endmembers are identically zero as shown in \textbf{Fig. \ref{fig:samson_abun_correct}}.

\subsection{Denoising}


We test the robustness of SCA to noise by augmenting Samson dataset with zero-mean white Gaussian noise. The variance of the noise is altered over runs to account for different Signal to Noise Ratio (SNR) \wrt the data. \textbf{Table \ref{tab:noise-metric}} shows the two error metrics for SNR ranging from 100 dB to 20 dB where the minimum volume penalization term $\lambda$ has to be increased to achieve similar error scales as obtained in a noiseless setting. Given that a user might not have the ground truth end-members at hand, tuning this hyper-parameter (penalty) can be done by visually observing the $(K-1)$ simplex on the abundances. The hyper-parameter values are altered until a three dimensional scatter plot of the abundances forms a perfect $(K-1)$ simplex. \textit{This also serves as an additional qualitative process to ensure that the end-members are extracted correctly from noisy data.} For SCA's low-rank approximation to extract the correct endmembers, in a noisy dataset, it is assumed that the SNR is such that the noise spectra itself is not high energy. In other words, the noise spectra is lower energy compared to all of the desired endmember spectra energies. 

\begin{table}[ht]
    \centering
    \vspace{-0.5em}
    \begin{tabular}{c|ccccc}
    \toprule
        SNR (dB) & 100 & 50 & 40 & 30 & 20 \\ \midrule
        $\lambda$ & 0.05 & 0.1 & 0.5 & 1.0 & 10.0 \\
        RMSE(A) (10$^{-3}$) & 3.48$\pm$0.7 & 5.37$\pm$0.9 & 13.6$\pm$0.1 & 21.8$\pm$0.1 & 32.2$\pm$0.1 \\
        SAD(E) (10$^{-4}$) & 5.68$\pm$0.1 & 8.96$\pm$0.2 & 17.0$\pm$0.0 & 
        25.8$\pm$0.0 & 40.3$\pm$0.0 \\
    \bottomrule
    \end{tabular}
    \caption{Optimal Values of $\lambda$ used to reach same error metrics, showing a linear dependency of $\lambda$ on SNR.}
    \label{tab:noise-metric}
    \vspace{-3mm}
\end{table}

\subsection{Effect of Outliers}

We now demonstrate that the self-correcting property of SCA due to a bi-orthogonal representation renders robustness against outliers. Here we differentiate outlier from noise since the former exhibits itself as a separate spectra compared to the latter which perturbs all spectra. An outlier in HSI is a data point that: \textbf{1)} although in the hyper-plane of the $(K-1)$ simplex lies outside the simplex or \textbf{2)} lies out of the hyper-plane of the simplex. In the following numerical experiment, we augment the Samson dataset with outliers that are a combination of both the aforementioned sub-categories. 
\begin{table}[ht]
    \centering
    \vspace{-0.5em}
    \begin{tabular}{c|ccccc}
    \toprule
        \#Outlier & 5 & 10 & 20 & 50 & 100 \\ \midrule
        RMSE(A) (10$^{-5}$) & 1.25$\pm$0.2 & 1.47$\pm$0.2 & 1.69$\pm$0.1 & 2.08$\pm$0.1 & 2.28$\pm$0.1 \\
        SAD(E) (10$^{-4}$) & 1.71$\pm$0.1 & 1.99$\pm$0.1 & 2.42$\pm$0.2 & 
        2.71$\pm$0.2 & 3.13$\pm$0.2 \\
    \bottomrule
    \end{tabular}
    \caption{Error metrics in the presence of outliers using Samson dataset. An over-specified $K+O=4$ was used instead $K=3$ to serve as an extra endmember capturing outliers. The variance in errors due to the presence of outliers is at the scale of the error itself.}
    \label{tab:outlier-metric}
    \vspace{-2mm}
\end{table}

As a general strategy, we over-specify the number of end-members and given SCA's self-correcting property the outliers occupy the over-specified endmember location. The outliers were generated from a uniform random distribution. \textbf{Table \ref{tab:outlier-metric}} presents five different cases with $K=4$ and the number of outliers varied from 5 to 100 demonstrating the robustness of SCA to outliers. If the dataset contains only outliers, the $\lambda = 0.001$ hyper-parameter remains the same as in a noiseless setting. This validates our strategy for treating outliers as additional endmembers which differs from handling noisy data.

\section{Runtime and Loss Profile}

\textbf{Table \ref{tab:runtime}} shows a runtime comparison for Jasper and Urban datasets with the exception of Samson for which corresponding numbers are not reported by other works. 

\begin{table}[ht]
    \centering
    \begin{tabular}{c|ccc|c}
    \toprule
        Method & DAEN \cite{su2019daen} & DCAE \cite{khajehrayeni2020hyperspectral} & Endnet \cite{ozkan2018endnet} & SCA \\ \midrule
        Jasper & 165 & 110 & 855 & 400 \\
        Urban & 870 & 500 & 914 & 1000 \\ \bottomrule
    \end{tabular}
    \caption{Runtime (secs) across neural HSI models. The first three methods require special initialization with VCA, hence lower training times.}
    \label{tab:runtime}
    \vspace{-2em}
\end{table}

\begin{figure}[ht]
    \centering
    \includegraphics[width=0.5\linewidth]{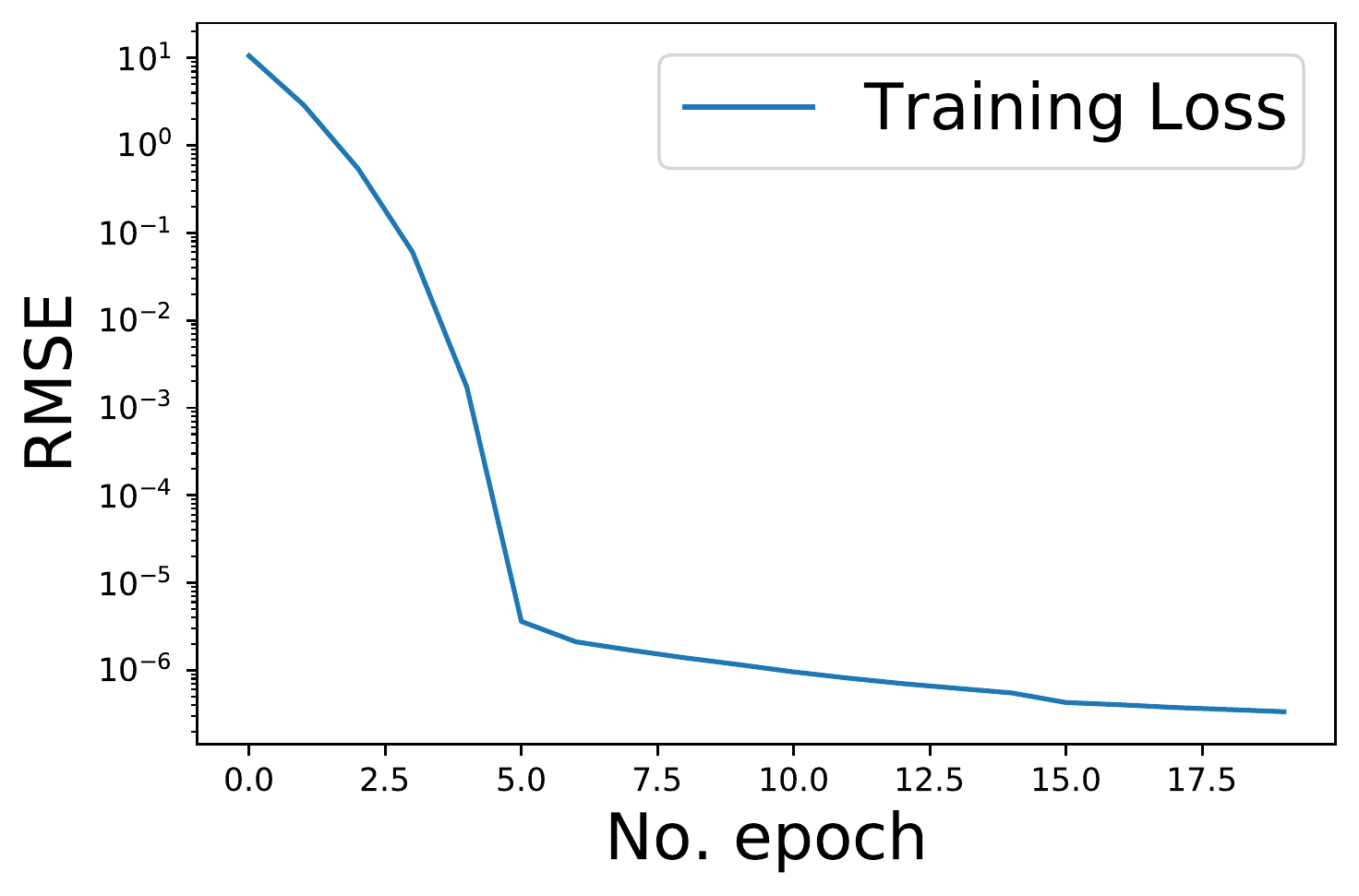}
    \vspace{-1em}
    \caption{Loss Profile for Samson dataset using SCA. Since we do not have any validation split as explained in Section \ref{sec:trval}, we only plot the training loss.}
    \vspace{-1em}
    \label{fig:epoch_samson}
\end{figure}

The network loss profile for Samson dataset is shown in \textbf{Fig. \ref{fig:epoch_samson}}. All our network runs across three datasets are done for $20$ epochs. As mentioned earlier, we do not perform any training-validation split so that the span of the dataset remains unchanged resulting in a deterministic tail energy bound.The total number of trainable parameters for an LMM (Eq. \ref{eq:lmm}) is $K(N+F)$. SCA has only $2FK$ parameters which is strictly less than $K(N+F)$ because for all the datasets $N \gg F$. Thus, our network cannot over-fit while approximating $E,A$ to jointly satisfy the LMM formulation.

\section{Conclusion}

We present a Self-Correcting Autoencoder SCA for HSI unmixing to extract the endmember spectra and abundances. The network formulation relies upon a bi-orthogonal representation where the identified endmembers are bi-orthogonal to an extracted dual spanning the top rank-$K$ space of the input data. We also provide tail energy bounds for the extracted representation following Eckart-Young-Mirsky theorem that can be computationally verified once the network converges. SCA network parameters are dictated by $2FK$, independent of the number of samples. 
The self-correcting property of SCA ensures that the endmembers are extracted correctly even if an over-specified $K$ is prescribed. Our numerical results on Samson, Jasper, and Urban datasets demonstrate that SCA error metrics are substantially better than the state of art methods with error metrics at scale $10^{-5}$ compared to previously reported $10^{-2}$. We also demonstrate the robustness of SCA to noise and outliers. 

\bibliographystyle{ACM-Reference-Format}
\bibliography{hsi-arxiv}

\appendix

\section{Low Rank Approximation}\label{app:low}

We also demonstrate the robustness of SCA-Net for an under-specified number of end-members $K-U$. Here, we consider the Jasper dataset where the number of ground-truth endmembers are known to be 4. The following numerical experiment considers extracting only 3 endmembers for testing purposes.
\begin{figure}[ht]
    \centering
    \includegraphics[width=0.95\linewidth]{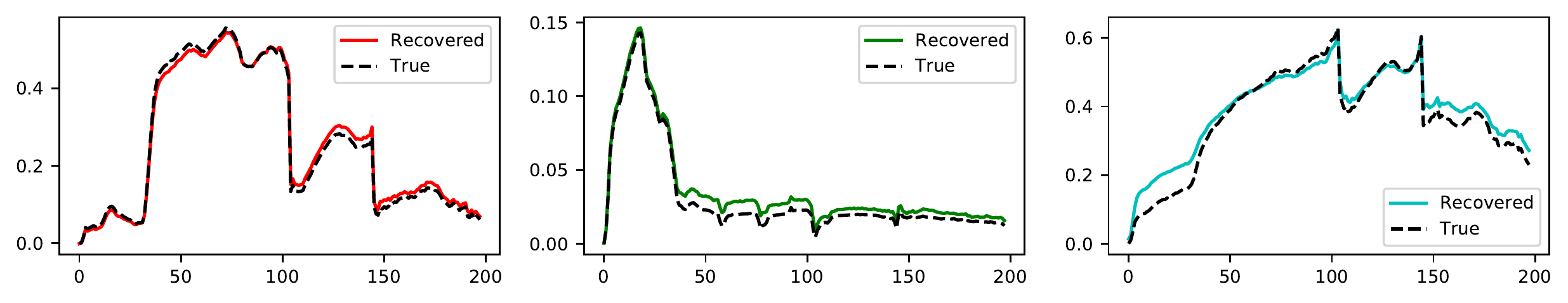}
    \caption{Extracted (solid line) and ground-truth (dashed-line) under-specified endmember spectra for Jasper dataset.}
    \label{fig:jasper_spec_low}
\end{figure}

\begin{figure}[ht]
    \centering
    \includegraphics[width=0.5\linewidth]{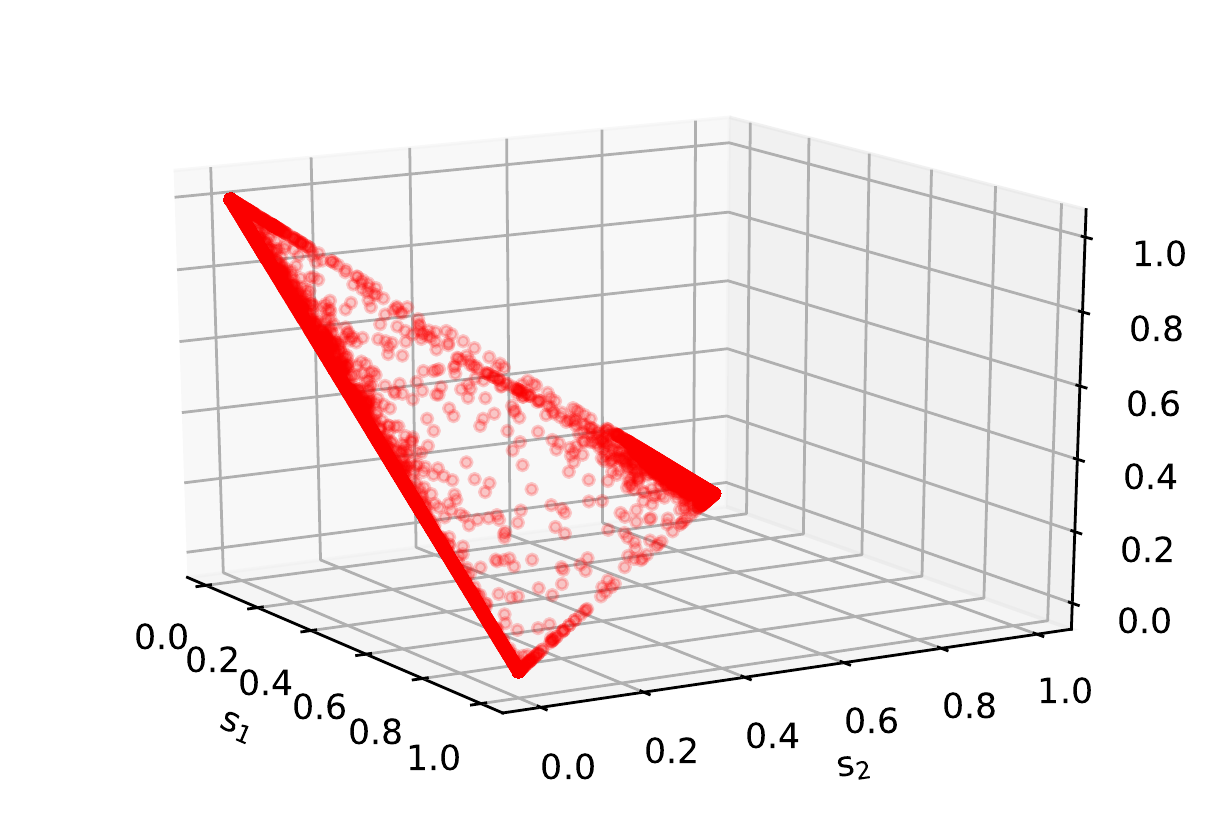}
    \caption{2-Simplex for the under-specified Jasper test case.}
    \label{fig:japser_simplex_low}
\end{figure}

 Fig. \ref{fig:jasper_spec_low} shows the extracted endmembers corresponding to the three highest energy spectra. Note that even under this incorrect specification the 2-simplex is still formed correctly when SCA-Net converges, as shown in Fig. \ref{fig:japser_simplex_low}. This serves as a numerical verification that the choice of our non-linear activation function in Section \ref{subsec:act} is correct. 

\section{Additional Results}\label{app:simplex}

\begin{figure}[ht]
    \centering
    \includegraphics[width=0.7\linewidth]{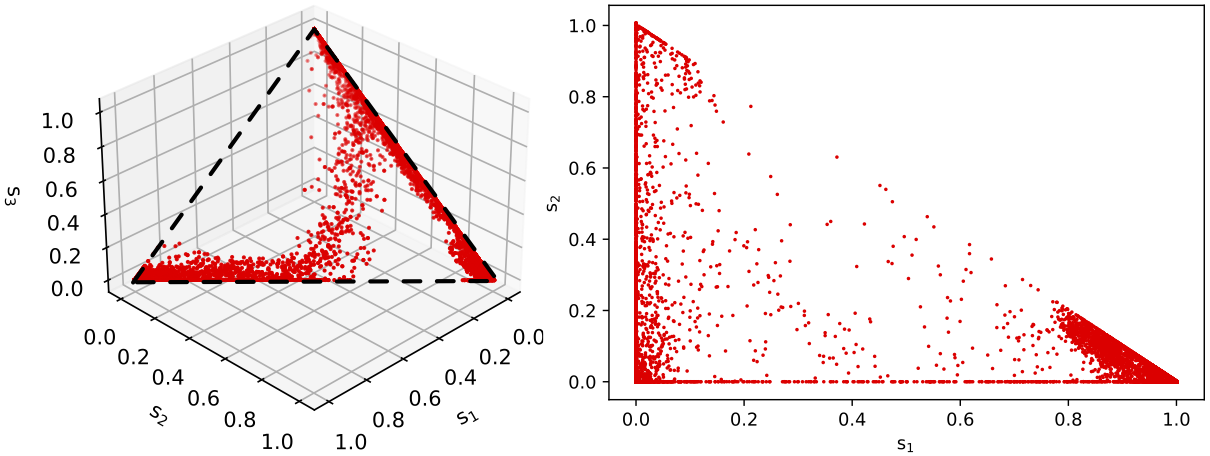}
    \caption{SCA-Net extracted 2-Simplex for Samson dataset}
    \label{fig:Samson_simplex_3d}
\end{figure}
In this section, we provide an addendum to the HSI results in the main text. Fig \ref{fig:Samson_simplex_3d} shows the recovered 2-simplex with the vertices representing the end-members for the Samson dataset. Since this dataset considers only three end-members this 2-simplex is an equilateral triangle satisfying the $\sum_{k} a_{k} = \mathbf{1}$ visually shown using a 3D scatter plot in Fig. \ref{fig:Samson_simplex_3d} (left). As expected, Fig. \ref{fig:Samson_simplex_3d} (right) shows a 2D projection as a right angled isosceles triangle. This serves as a means to identify the hyper-parameter $\lambda$ wherein the endmember spectra are recovered adequately. 
\begin{figure}[ht]
    \centering
    \includegraphics[width=0.8\linewidth]{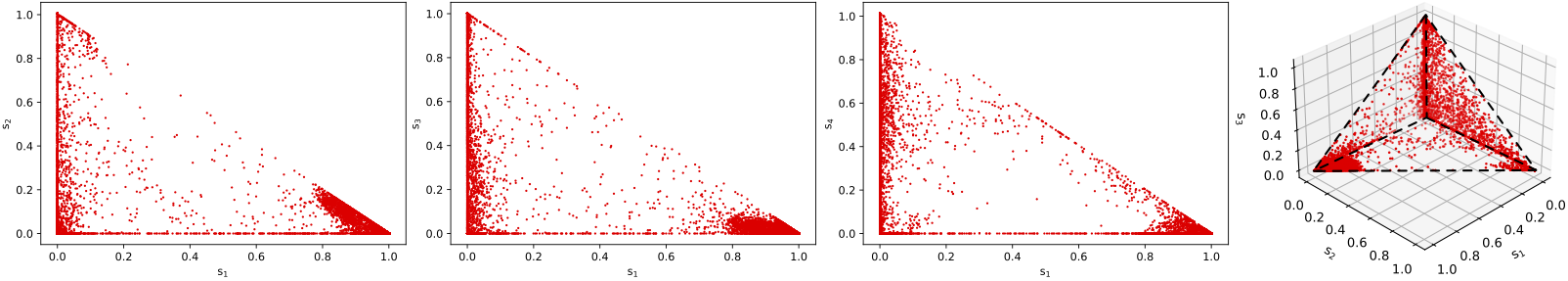}
    \caption{SCA-Net extracted 3-Simplex for Jasper dataset}
    \label{fig:Jasper_simplex_3d}
\end{figure}

Fig. \ref{fig:Jasper_simplex_3d} shows a similar plot for the recovered 3-simplex for the Jasper dataset. Since the number of end-members are $>3$, the 2D-projection serves as a convenient visual aid to ratify the recovered end-member spectra and for tuning  the hyper-parameter $\lambda$ in a noisy dataset.

\end{document}